\documentclass[11pt]{article}

\usepackage[preprint]{acl}

\usepackage{times}
\usepackage{latexsym}

\usepackage[T1]{fontenc}

\usepackage[utf8]{inputenc}

\usepackage{microtype}

\usepackage{inconsolata}

\usepackage{graphicx}

\usepackage{hyperref}
\usepackage{url}
\usepackage{xcolor} 
\usepackage[most]{tcolorbox}
\usepackage{colortbl}
\usepackage{color}

\usepackage{multirow}
\usepackage{makecell}
\usepackage{booktabs}
\usepackage{subcaption}
\usepackage{wrapfig}
\usepackage{bbm}
\usepackage{amssymb}
\usepackage{diagbox}

\definecolor{RedOrange}{cmyk}{0,0.77,0.87,0}
\definecolor{BlueGreen}{cmyk}{0.85,0,0.33,0}

\title{Learning to Refine: Self-Refinement of Parallel Reasoning in LLMs}

\author{
  \textbf{Qibin Wang\textsuperscript{1,2}},
  \textbf{Pu Zhao\textsuperscript{1}\thanks{Corresponding author.}},
  \textbf{Shaohan Huang\textsuperscript{1}},
  \textbf{Fangkai Yang\textsuperscript{1}},
  \textbf{Lu Wang\textsuperscript{1}},
\\
  \textbf{Furu Wei\textsuperscript{1}},
  \textbf{Qingwei Lin\textsuperscript{1}},
  \textbf{Saravan Rajmohan\textsuperscript{1}},
  \textbf{Dongmei Zhang\textsuperscript{1}}
\\
\\
  \textsuperscript{1}Microsoft,
  \textsuperscript{2}Peking University
\\
  \small{
    \texttt{wangqibin@stu.pku.edu.cn},
    \texttt{\{Pu.Zhao, shaohanh, fangkai.yang, wlu, fuwei, qlin, sarr1, dongmeiz\}@microsoft.com}
  }
}

\begin{document}
\maketitle
\begin{abstract}
Test-time scaling (TTS) has gained widespread attention for enhancing LLM reasoning. Existing approaches such as Best-of-N and majority voting are limited as their performance depends on the quality of candidate responses, making them unable to produce a correct solution when all candidates are incorrect. 
Parallel self-refinement, generating multiple candidates and synthesizing a refined answer conditioned on them, offers a promising alternative, but the underlying mechanism driving its effectiveness remains obscure. 
To bridge this gap in understanding, we introduce a new metric, the Refinement Gap, designed to quantify the relative improvement of self-refinement beyond majority voting.
We show that the Refinement Gap exhibits a clear scaling trend with model size and is only weakly correlated with the base capability.
Based on this discovery, we propose Generative Self-Refinement (GSR), a parallel test-time scaling framework that transfers the refinement policy from larger teacher models with higher refinement gap into smaller students. Crucially, GSR jointly trains a single model to generate strong candidates and refine a better final answer based on these candidates.
Experimental results demonstrate that our method achieves state-of-the-art performance across five mathematical benchmarks over other parallel aggregation methods, while the learned refinement skill transfers across multiple model scales and families and exhibits robust generalization to an out-of-distribution domain.
\end{abstract}

\section{Introduction}
Test-time scaling (TTS) \citep{brown2024largelanguagemonkeysscaling, snell2024scalingllmtesttimecompute} has become a dominant strategy, in which model performance can be consistently improved by allocating additional compute at inference time.
A common strategy is majority voting \citep{wang2023selfconsistency}, which leverages the principle of self consistency to improve performance, identifying the most frequent answer from multiple reasoning paths. Best-of-N (BoN) approach takes this process further by introducing an external verifier, typically a Reward Model (RM), to score, rank and select the best response from candidates \citep{Stiennon2020summarizewithhumanfeedback,irvine2023rewardingchatbotsrealworldengagement, song2024goodbadgreedyevaluation}. 
However, these prevailing strategies are inherently bounded by the quality of the set of candidates \citep{khairi2025makingtakingbestn}. They are limited to produce a solution that transcends the quality of candidate proposals, which becomes particularly problematic when all candidates are flawed \citep{zhang2025cotbasedsynthesizerenhancingllm}. 

\begin{figure}[th]
    \centering
    \includegraphics[width=1.0\linewidth]{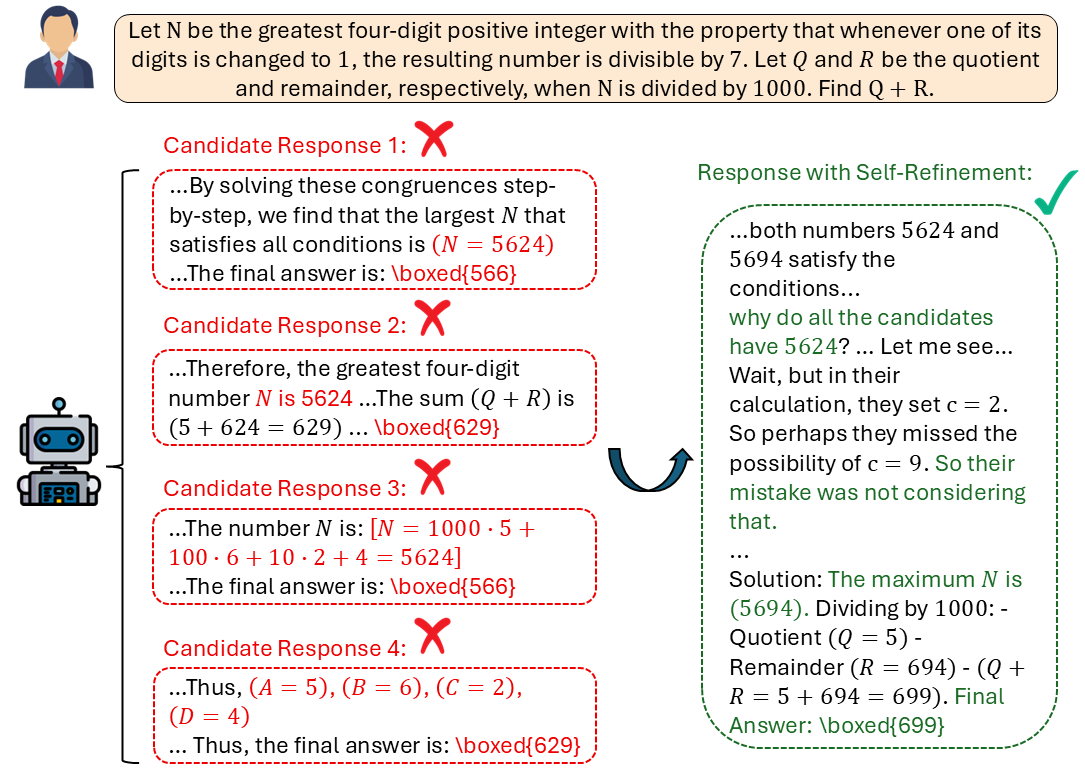} 
    \caption{
    Case study of Qwen2.5-7B-Instruct after training. Even provided with four incorrect candidates, it can still reference them, diagnose the flaws, and finally construct a correct answer. We provide a more detailed case in Appendix \ref{sec:case_study}.
    }
    \label{fig:example}
\end{figure}

To address this limitation, several refinement methods \citep{madaan2023selfrefineiterativerefinementselffeedback, kim2023languagemodelssolvecomputer, paul2024refinerreasoningfeedbackintermediate}, where a model critiques and improves its own output, provide a promising alternative. These works have often focused on sequential refinement, where the model iteratively refines its answers. However, \citet{huang2024largelanguagemodelsselfcorrect} finds that LLMs struggle to self-refine their own reasoning without external feedback and in some cases can even degrade performance, highlighting the limitations of intrinsic sequential self‑refinement. 
Consequently, recent research has shifted towards parallel aggregation methods \citep{vernikos-etal-2024-small, zhang2025cotbasedsynthesizerenhancingllm, li2025llmsgeneratebetteranswer}, attempting to synthesize a correct solution from multiple generated candidate solutions.
While these approaches leverage candidate diversity, the fundamental mechanism behind the success remains obscure.  It is  unclear whether the improvement in parallel refinement is driven by better candidate coverage, enhanced reasoning abilities, or emergent interaction effects among multiple samples.
This gap in understanding motivates us to develop a metric for evaluating intrinsic refinement ability. Therefore, we propose 
\textbf{the Refinement Gap}, which quantifies the improvement of refinement beyond majority voting. 
Based on this metric, we reveal a surprising systematic scaling trend: \textbf{the refinement capability is primarily governed by model scale, with minimal correlation to domain-specific capabilities.}

\begin{figure*}[ht!]
\centering
\includegraphics[width=1.0\linewidth]{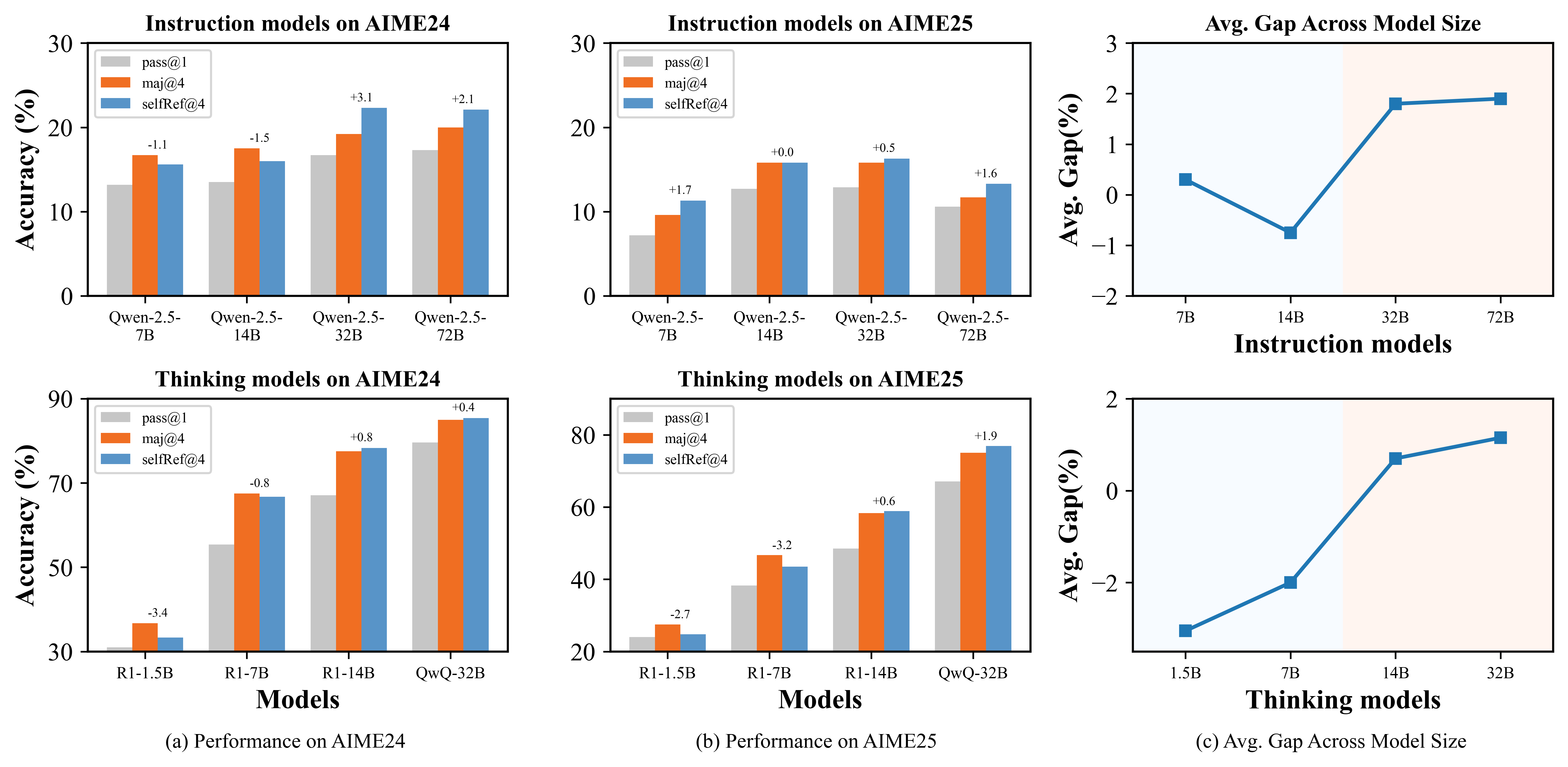} 
\caption{
A comprehensive study of refinement gap. (\textbf{\textit{Left}}) Accuracy of instruction models on AIME24/AIME25 under pass@1, majority voting (maj@4), and parallel self-refinement (selfRef@4). Numbers denote $\Delta=\text{selfRef@4}-\text{maj@4}$. (\textbf{\textit{Middle}}) Accuracy of thinking models on AIME24/AIME25. (\textbf{\textit{Right}}) Average $\overline{\Delta}$ across AIME24/AIME25 versus model size. 
Shaded regions highlight low and high gain according to $\overline{\Delta}$.
}
\label{fig:refinement_gap}
\end{figure*}

To address this gap, we introduce \textbf{Generative Self-Refinement (GSR)}, a training-enabled parallel test-time scaling framework designed to explicitly activate self-refinement in smaller models. 
GSR distills a refinement policy from high performance teacher models with a large refinement gap into smaller student models, enabling them to generate improved solutions beyond simple selection. 
Crucially, we employ \textbf{a hybrid training pipeline} that jointly optimizes for direct solution generation and refinement.
This hybrid training pipeline ensures a synergy where strong base reasoning improves the refinement process, while refinement feedback enhances reasoning robustness.
As illustrated in Figure~\ref{fig:example}, we empirically demonstrate that GSR enables small models to transcend the limits of selection, successfully recovering correct solutions even in the difficult scenarios where all candidates are incorrect.

Our key contributions are as follows:

\begin{itemize}
    \item We introduce the Refinement Gap, a metric that systematically quantifies the intrinsic gain of self-refinement beyond majority voting. This metric reveals a scaling trend in refinement ability across model sizes. 

    \item We propose Generative Self-Refinement (GSR),  a training-enabled parallel test-time scaling framework that transfers the refinement policy from larger teacher models to smaller models.

    \item We demonstrate that our method achieves state-of-the-art results empirically on five challenging mathematical benchmarks over other parallel aggregation methods and refinement skill is robust under challenging candidate conditions.
    
\end{itemize}

\section{Scaling Trend of Refinement Gap}
\label{sec:scaling}
Recent works \citep{vernikos-etal-2024-small, zhang2025cotbasedsynthesizerenhancingllm, li2025llmsgeneratebetteranswer} have considered parallel self-refinement as an effective method of test-time scaling.
However, the underlying source of parallel self-refinement gain is still underexplored.
In particular, even a simple self-consistency aggregation (majority voting) over the same candidate solutions can substantially improve accuracy, making it unclear how much improvement should be attributed to refinement procedure.
This motivates our first research question: (\textbf{RQ1}) \textit{Do we really need parallel self-refinement?}

To quantify the marginal gain of parallel self-refinement, we define the \textbf{Refinement Gap} as 
\begin{equation}
\Delta(\mathcal{M},\mathcal{D},k) = {\text{selfRef@k}}(\mathcal{M},\mathcal{D}) - {\text{maj@k}}(\mathcal{M},\mathcal{D})
\end{equation}
for a given model $\mathcal{M}$, dataset $\mathcal{D}$ and candidate budget $k$.
We use majority voting as \textit{a natural gold baseline} because it is a deterministic function operator over the same solutions and is expected to improve monotonically with pass@1 under the independence assumptions \citep{kuncheva2014combining}.
We conduct a systematic study across model scales from 1.5B to 72B across multiple model families including instruction Qwen-2.5-Instruct (Qwen-2.5) and thinking DeepSeek-R1-Distill-Qwen/QwQ (R1/QwQ), on the competitive mathematics benchmarks AIME24/AIME25, to enable a comprehensive analysis of the effectiveness of parallel self-refinement.

As illustrated in Figure \ref{fig:refinement_gap}, we summarize the key observations as follows:
\textbf{(1) Parallel self-refinement consistently raises the ceiling beyond the base capability measured by pass@1 across all models and benchmarks evaluated.}
\textbf{(2) The Refinement Gap $\Delta$ exhibits a clear scaling trend across instruction and thinking models.} 
Further, we empirically observe \textit{a regime shift} around 14B. Both instruction and thinking models $\geq$ 14B tend to achieve a higher average refinement gain $\overline{\Delta}$ than smaller models. We visualize these low and high gain regimes with different shaded regions in Figure~\ref{fig:refinement_gap}.
\textbf{(3) The Refinement Gap $\Delta$ is weakly correlated with the base capability of the models.} For instance, on AIME24, DeepSeek-R1-Distill-Qwen-7B attains a higher base capability (54.4\%) yet has $\overline{\Delta}=-0.8$, while Qwen2.5-32B-Instruct attains a larger $\overline{\Delta}=3.1$ despite a much lower pass@1(18.6\%).

\begin{figure*}[th!]
\centering
\includegraphics[width=1.0\textwidth]{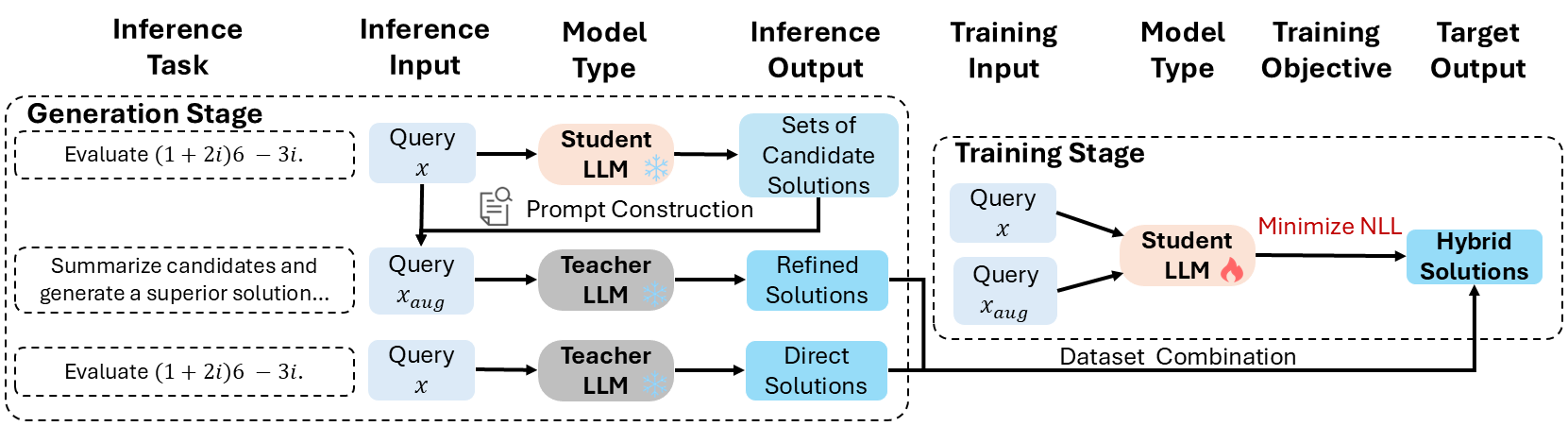} 
\caption{
An overview of the hybrid training pipeline, which consists of a data generation stage (\textbf{\textit{Left}}) followed by a supervised fine-tuning (SFT) stage (\textbf{\textit{Right}}). We use a teacher model to construct hybrid dataset and then train the student model on dual tasks.
}
\label{fig:trainingpipeline}
\end{figure*}

\section{Methodology}
\subsection{Overview}
Based on the analysis in Section \ref{sec:scaling}, we identify that the refinement gap $\Delta$ exhibits a clear scaling trend. Empirically, smaller models often reside in a low gain regime, whereas larger models consistently exhibit a much higher refinement gap.
Crucially, this capability cannot be reliably elicited in smaller models through prompting, motivating our hypothesis that \textbf{self-refinement is a learnable inference strategy that must be explicitly activated via training}.

To bridge this gap, we propose \textbf{Generative Self-Refinement (GSR)}, a parallel test-time scaling framework designed to transfer the self-refinement capability from larger models with higher gain to smaller models. At test time, GSR follows a generate-then-refine procedure, where a unified model samples $k$ candidate solutions and then synthesizes a self-refined final answer.

\subsection{Data Construction Pipeline}
Guided by Section~\ref{sec:scaling}, we select large teacher models with high refinement gap to ensure reliable refinement supervision for small models. 
To this end, we construct refinement triples through the following steps as illustrated  in Figure~\ref{fig:trainingpipeline}. For each problem $x$, we first sample $k$ candidate solutions $\mathcal{C}_k=\{c_i\}_{i=1}^k$ from the student itself to match the test-time candidate distribution. We then build an augmented input $x_{aug}=\text{Aug}(x,\mathcal{C}_k)$ by concatenating the question with the enumerated candidates (see Appendix \ref{sec:prompt} and \ref{sec:analysis_input_burden} for more details). Subsequently, we query the teacher to produce a refined target solution $y^T_{ref}$. This procedure yields the refinement dataset $\mathcal{D}_{\text{ref}}=\{(x^{(i)}, \mathcal{C}_k^{(i)}, {y^{T}_{\text{ref}}}^{(i)})\}_{i=1}^N$. Unlike vanilla distillation, this supervision teaches how to refine the student’s own candidates, rather than only imitating direct answers. 
Notably, to ensure the student learns synthesis beyond candidate responses selection, we intentionally retain hard instances where the candidate pool is of low quality and highly inconsistent. Specifically, we emphasize the cases with $N_c=0$ (i.e., all candidate responses are incorrect) as well as other highly inconsistent scenarios, where voting or ranking is not effective. These critical cases provide direct supervision for constructing a correct solution by integrating partial insights and correcting errors from imperfect candidates.

In addition to the refinement dataset $\mathcal{D}_{\text{ref}}$, we collect direct solutions to maintain and strengthen the base capability of student. Concretely, we query the teacher with the original question $x$ to obtain a direct target solution $y^{T}_{\text{dir}}$ and forms a direct-solving dataset $\mathcal{D}_{\text{dir}}=\{(x^{(i)},  {y^{T}_{\text{dir}}}^{(i)})\}_{i=1}^N$. 
This component is crucial because the candidate quality sets the performance ceiling for refinement and stabilizes the joint optimization in Section~\ref{sec:hybrid_train_pipeline}.

\subsection{Hybrid Training Objective}
\label{sec:hybrid_train_pipeline}

We optimize a single student model $\mathcal{S}_\theta$ to acquire two complementary skills, \textbf{direct-solving} for generating strong candidates and \textbf{refinement} for synthesizing a better final solution given these candidates, aiming for a higher performance ceiling. As shown in Figure~\ref{fig:trainingpipeline}, we optimize model $\mathcal{S}_\theta$ on the direct-solving dataset $\mathcal{D}_{\text{dir}}$ and the refinement dataset $\mathcal{D}_{\text{ref}}$. Formally, we minimize the negative log-likelihood (NLL) objectives as follows:
\begin{equation}
    \begin{aligned}
&\mathcal{L}_{\text{dir}}(\theta) = - \mathbb{E}_{(x,y_{\text{dir}}^T)\sim \mathcal{D}_{\text{dir}}}        [\log \mathcal{S}_\theta(y^{T}_{\text{dir}}|x)] \\
&\mathcal{L}_{\text{ref}}(\theta) = - \mathbb{E}_{(x,\mathcal{C}_k,y_{\text{ref}}^T)\sim \mathcal{D}_{\text{ref}}}        [\log \mathcal{S}_\theta(y^{T}_{\text{ref}}|x_{\text{aug}})] \\
    \end{aligned}
\end{equation}
Then, we optimize a weighted combination:
\begin{equation}
        \mathcal{L}(\theta) = \lambda \mathcal{L}_{\text{ref}}(\theta) + (1-\lambda) \mathcal{L}_{\text{dir}}(\theta)
\end{equation}
where $\lambda$ controls the trade-off between learning refinement behavior and improving base capability. For simplicity, we set $\lambda=0.5$.

Joint optimization for hybrid objectives is essential to the success of GSR. Training only the refinement objective tends to be bottlenecked by candidate quality, which in turn limits the headroom. In contrast, training only the direct-solving objective improves the base capability but does not reliably teach the model to leverage imperfect candidates, resulting in a low or unstable refinement gap $\Delta$.  
By jointly optimizing direct solving and refinement, the model preserves strong base capability while learning the refinement strategy, getting complementary gains in both candidate quality and refinement effectiveness.
At test time, the unified model $\mathcal{S}_\theta$ performs generate-then-refine by sampling $k$ candidates and synthesizing a refined answer \textbf{without any external module}.

\section{Experiments}

\begin{table*}[ht!]
\small
\centering
  \caption{
  Comprehensive performance evaluation on five mathematical benchmarks. We report the results and the Refinement Gap $\Delta$ (selfRef - maj) of our GSR-7B and base Qwen2.5-7B-Instruct for reference. Additionally, we report the pass@1 metric for two leading models from \citet{yang2025reasonfluxhierarchicalllmreasoning}.
 }
  \begin{tabular}{ll|ccccc|c}
    \toprule
    \textbf{Method} & \textbf{Aggregation Model} & \textbf{AIME24} & \textbf{AIME25} & \textbf{AMC22-23} & \textbf{MATH500} &  \textbf{Olympiad} & \textbf{Avg.} \\
    \midrule
    \multicolumn{8}{c}{Larger Models} \\
    \midrule
    QwQ-Preview & - & 50.0 & & & 90.6 & 61.2 \\
    o1-mini & - & 56.7 & & & 90.0 &  \\

    \midrule
    \multicolumn{8}{c}{Qwen2.5-7B-Instruct (\textit{Base})} \\
    \midrule
    pass@1 & - & 13.2 & 7.2 & 43.6 & 75.9 & 39.5 & 35.9\\
    maj@4 & - & 16.7 & 9.6 & 47.7  & 79.4 & 43.5 & 39.4\\
    BoN@4 & SkyworkRM-27B &17.1 & 9.2 & 48.0  & 78.6 & 43.0 & 39.2\\
    selfRef@4 & - & 15.6 & 11.3 & 47.5  & 78.7 & 43.8 & 39.4\\
    $\Delta$ &  &  \cellcolor{gray!10} {\color{BlueGreen} -1.1} & \cellcolor{gray!10} {\color{RedOrange} +1.7} & \cellcolor{gray!10} {\color{BlueGreen} -0.2} & \cellcolor{gray!10} {\color{BlueGreen} -0.7} & \cellcolor{gray!10} {\color{RedOrange} + 0.3} & \cellcolor{gray!10}{\color{BlueGreen} \textbf{$\pm$0.0}}  \\

    \midrule
    \multicolumn{8}{c}{GSR-7B (\textit{After Training})} \\
    \midrule
    
     pass@1 & - & 50.1 & 37.8 & 78.5 & 90.6 & 64.4 & 64.3 \\
     maj@4 & - & 60.0 & 46.7 & 84.6 & 92.8 & 68.3 & 70.5\\
     maj@5 & - & 60.7 & 48.3 & 84.9 & 93.1 & 68.8 & 71.1 \\

     BoN@4 & RRM-7B & 60.4 & 46.3 & 80.4 & 90.6 & 64.6 & 68.5 \\
     
     BoN@4 & RM-R1-7B & 62.1 & 45.4 & 84.6 & \textbf{93.5} & 69.1 & 70.9\\
     BoN@4 & SkyworkRM-27B & 58.3 & 46.7 & 81.6 & 92.5 & 67.6 & 69.3\\
     BoN@5 & SkyworkRM-27B & 59.2 & 46.3 & 82.8 & 92.7 & 67.7 & 69.7\\
     Ref@4 & Synthesizer-8B & 60.2 & 44.0 & 84.0 & 92.0 & 67.7 & 69.6\\
     selfRef@4 & - & \textbf{66.0} & \textbf{51.7} & \textbf{85.7} & 93.4 & \textbf{71.0} & \textbf{73.6}\\

     $\Delta$ &  &  \cellcolor{gray!10} {\color{RedOrange} + 6.0} & \cellcolor{gray!10} {\color{RedOrange} + 5.0} & \cellcolor{gray!10} {\color{RedOrange} + 1.1} & \cellcolor{gray!10} {\color{RedOrange} + 0.6} & \cellcolor{gray!10} {\color{RedOrange} + 2.7} & \cellcolor{gray!10}{\color{RedOrange} \textbf{+3.1} } \\
  \bottomrule
\end{tabular}
  \label{tab:main_result}
\end{table*}

\subsection{Experimental Setup}
\paragraph{Dataset Curation}
We construct our training corpus from OpenMathReasoning \citep{moshkov2025aimo}, a large math dataset generated by Deepseek-R1 \citep{deepseekr1} and QwQ-32B\citep{qwq32b}. We retain 184k direct-solving instances and derive an additional 184k refinement instances, where each instance pairs the original query with four candidate responses. Details are provided in Appendix \ref{sec:dataset_curation}.

\paragraph{Training Settings}
We perform supervised fine-tuning (SFT) on Qwen2.5-7B-Instruct~\citep{qwen2.5} and denote the resulting model as \textbf{GSR-7B}. We present more training details in Appendix \ref{sec:training_settings}.

\paragraph{Aggregation Baselines}
To rigorously evaluate the efficacy, we compare with the following aggregation TTS baselines:
(1) \textbf{Majority Voting} \citep{wang2023selfconsistency}, a method selects the most consistent answer without any external modules. 
(2) \textbf{Skywork-Reward-Gemma-2-27B-v0.2} (SkyworkRM-27B) \citep{liu2024skyworkrewardbagtricksreward}, a SOTA scalar reward model which ranks the overall quality without explanations or reasoning. 
(3) \textbf{RM-R1-DeepSeek-Distilled-Qwen} \citep{chen2025rmr1rewardmodelingreasoning} (RM-R1) and \textbf{RRM} \citep{guo2025rewardreasoningmodel}, two concurrent approaches integrating reasoning capabilities into reward modeling, significantly surpassing conventional RM's performance. 
(4) \textbf{Synthesizer-8B-math} (Synthesizer-8B) \citep{zhang2025cotbasedsynthesizerenhancingllm},  an aggregation method  which synthesizes multiple candidates to produce a final answer, achieving SOTA performance on math domain.

\paragraph{Benchmarks}
For a comprehensive evaluation of mathematical performance, we evaluate all baselines across five challenging and most representative benchmarks \citep{hochlehnert2025soberlookprogresslanguage} in the mathematical domain: AIME24 \citep{aimo_validation_aime}, AIME25 \citep{aime2025}, AMC22 \& AMC23 \citep{aimo_validation_amc}, MATH500 \citep{hendrycks2021measuringmathematicalproblemsolving} and OlympiadBench \citep{he2024olympiadbenchchallengingbenchmarkpromoting}. 

\paragraph{Evaluation Settings}
For a direct and fair comparison, we report the average metrics maj@4 (majority voting), BoN@4 (Best-of-N) and Ref/SelfRef@4 (Refinement/Self-Refinement) on the same sets of four candidate responses ($k=4$).
To ensure a fair comparison of computational budget, we additionally report the average metrics for majority voting and non-generative BoN methods with five candidates ($k=5$).
We report the average metrics of 32 runs for AIME24 and AIME25 and 16 runs for the remaining. 
More details can be found in Appendix \ref{sec:evaluate}.

\subsection{Main Results}
\label{sec:main results}
We present a comprehensive evaluation of our method against advanced parallel test-time scaling baselines. The results, summarized across five challenging mathematical benchmarks, are presented in Table \ref{tab:main_result}.
A primary observation is the substantial performance shift after the training process. Compared to the base model, GSR-7B exhibits dramatic improvements across all standard metrics, particularly on the most challenging benchmarks. For instance, its pass@1 accuracy on AIME24 increases from a modest 13.2\% to 50.1\%, confirming the efficacy of our approach.

Our method shows significant improvement after post-training. Before training, the performance of selfRef@4 on the base model is comparable with majority voting.
After training, our method emerges as the \textbf{state-of-the-art (SOTA)} method, achieving 73.6\% average accuracy. It surpasses not only standard baselines such as majority voting but also more complex Best-of-N (BoN@4), as demonstrated on AIME24 (66.0\% vs. 62.1\%).  
\textbf{GSR-7B even achieves a higher average Refinement Gap $\Delta$ on AIME24/AIME25 (+5.5\%) than strong teacher model QwQ-32B (+1.15\%).}
We note that in the MATH500 dataset, selfRef@4 is marginally underperformed by BoN@4 (93.4\% vs. 93.5\%). 
We hypothesize that it is due to the specific dynamics of high-accuracy regimes (90.6\% on pass@1). In such a scenario, at least a correct answer is highly likely to be present in any set of candidate responses, making selection methods like Best-of-N particularly effective.


\begin{figure*}[th]
\centering

\begin{minipage}[t]{0.70\textwidth}
  \vspace{0pt} 
  \centering
  \includegraphics[width=\linewidth]{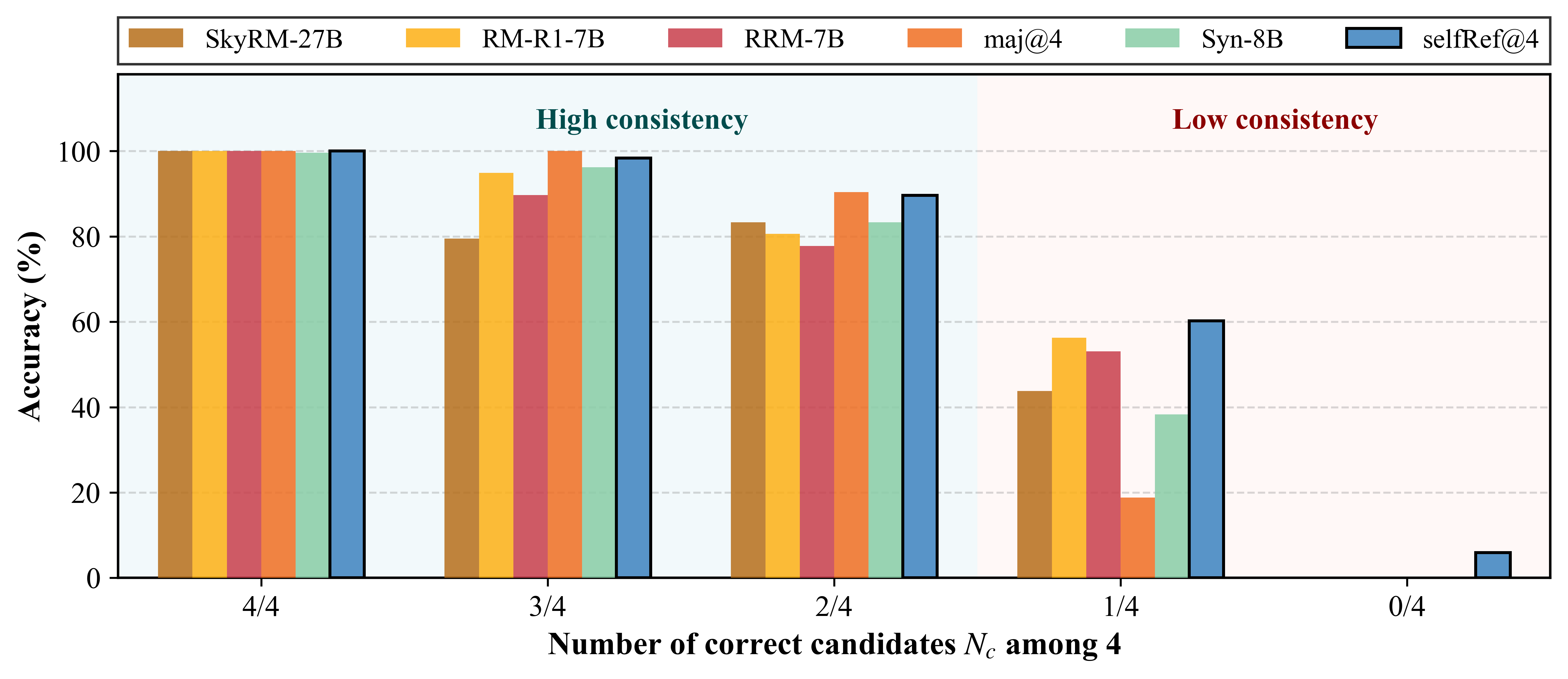}
  \makeatletter\def\@captype{figure}\makeatother
  \caption{Fine-grained analysis on the AIME24. We report accurate rates in percentage (\%) conditioned on the number of correct candidates ($N_c$).}
  \label{fig:baseline_correct_ratio_aime24}
\end{minipage}
\hspace{0.8em}
\begin{minipage}[t]{0.26\textwidth}
  \vspace{0pt} 

  \centering
  \includegraphics[width=\linewidth]{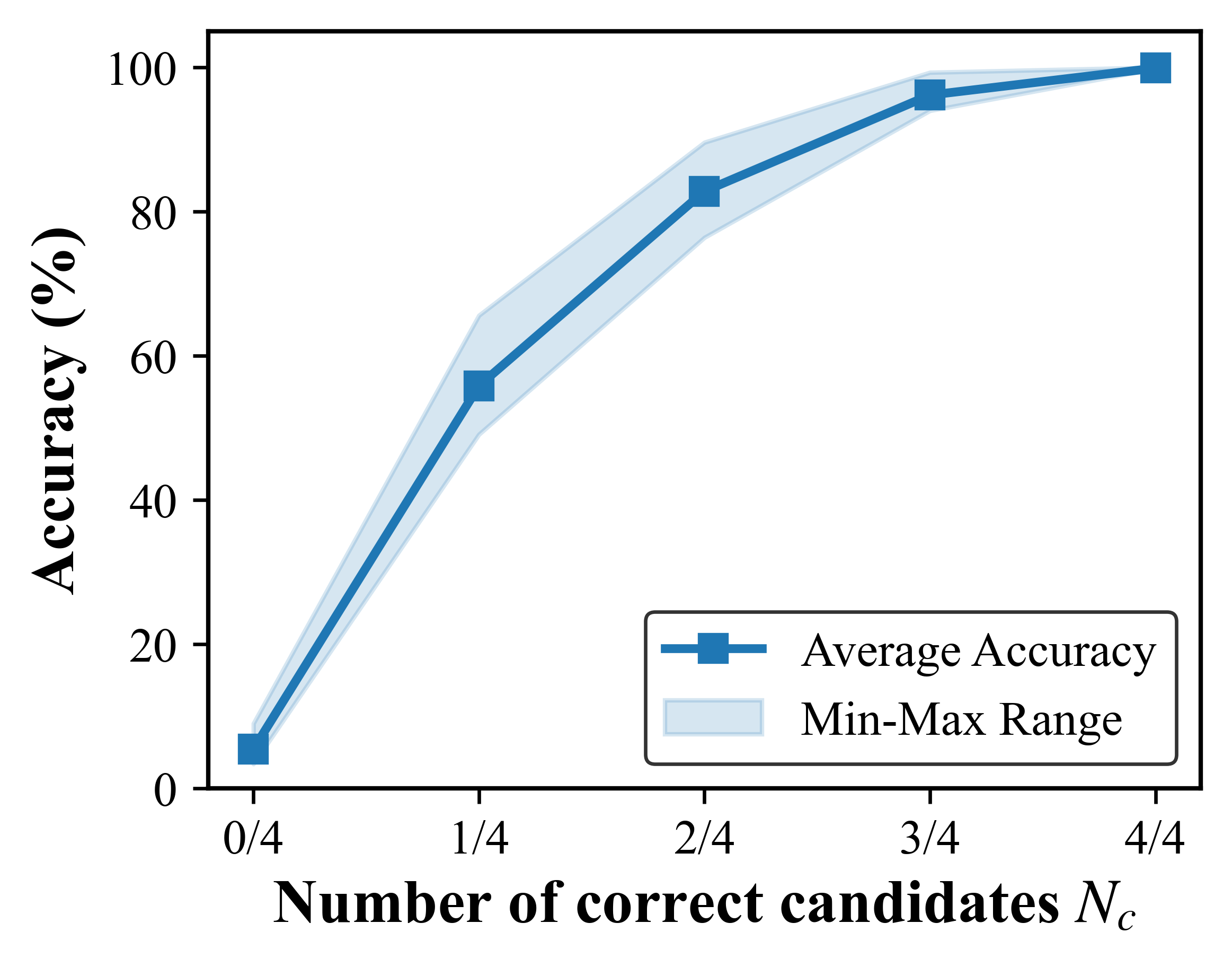}

  \makeatletter\def\@captype{figure}\makeatother
  \caption{Robustness analysis of our method across five diverse benchmarks. The full details are provided in Appendix \ref{sec:fine_grained_full_results} Table \ref{tab:full_correct_ratio}.}
  \label{fig:correct_ratio}
\end{minipage}

\end{figure*}

\subsection{Ablation Study on Training}
\label{sec:ablation dual-training}
In this section, we conduct an ablation study to answer: (\textbf{RQ2}) \textit{Does our hybrid training pipeline improve the upper bound of accuracy more than single mode training pipeline (direct-solving or refinement)?}
To ensure a fair comparison, we construct three distinct training datasets randomly sampled from the total dataset, all of which contain 20k samples and share an identical set of questions: 
(1) Direct-Solving (DS): 20k samples from $\mathcal{D}_{\text{dir}}$, also denoted as a vanilla knowledge distillation baseline.
(2) Refinement (Ref): 20k samples from $\mathcal{D}_{\text{ref}}$.
(3) Ours: a balanced mix of 10k from $\mathcal{D}_{\text{dir}}$ and 10k from $\mathcal{D}_{\text{ref}}$.

\begin{table}[ht]
\centering
\small 
\caption{Ablation study on different training strategies. }
\label{tab:ablation}
\begin{tabular}{lcccc}
\toprule
\multirow{2}{*}{\textbf{Method}}  & \multicolumn{2}{c}{\textbf{AIME24}} & \multicolumn{2}{c}{\textbf{AIME25}} \\
\cmidrule(lr){2-3} \cmidrule(lr){4-5}
&\textbf{pass@1} & \textbf{selfRef@4} & \textbf{pass@1} & \textbf{selfRef@4} \\
\midrule
baseline & 13.2 & 15.6 & 7.2 & 9.6\\
+\textit{DS} & \textbf{38.3} & 44.3 & \textbf{29.6} & \underline{33.2} \\
+\textit{Ref} & 26.4 & \underline{45.0} & 23.5 & 30.0 \\
\rowcolor{blue!6} +\textit{ours} & \underline{37.5} & \textbf{51.5} & \underline{29.0} & \textbf{36.5} \\

\bottomrule
\end{tabular}
\end{table}

The results presented in Table \ref{tab:ablation} lead to a clear conclusion. The model trained on a DS dataset achieves the highest pass@1 scores while \textbf{its' capacity for self-refinement is severely limited.}
To our surprise, even with training purely on the Ref dataset, the model still \textbf{achieves a higher pass@1 score than the base model.} However, this model excels at the self-refinement task but demonstrates \textbf{subpar pass@1 performance compared with two others.}
Our method strikes the optimal balance. It achieves the highest selfRef@4 scores across both benchmarks while maintaining a highly competitive pass@1 performance, confirming the necessity of the hybrid training pipeline.

\subsection{Fine-Grained Analysis on Self-Refinement}
\label{sec:fine-grained analysis}

Although the overall accuracy metrics are informative in Section \ref{sec:main results}, the underlying mechanism of our approach is still obscure. 
We conduct a more fine-grained conditional analysis across all methods when provided with imperfect or even entirely incorrect candidates. We report the average accurate rates conditioned on the number of correct candidates out of 4 ($N_c\in\{0,1,2,3,4\}$).

\paragraph{Comparison with Baselines} As shown in Figure~\ref{fig:baseline_correct_ratio_aime24},  when a clear majority of candidates are correct ($N_c \geq 3$), all methods perform exceptionally well, often achieving near perfect accuracy. However, our methods begin to differentiate in more ambiguous cases ($N_c<3$). When only one candidate is correct, our selfRef@4 achieves an impressive 60.2\% accuracy, substantially outperforming majority voting (maj@4 at 18.8\%) and all BoN@4 variants. Crucially, even when all candidates are incorrect ($N_c=0$), \textbf{our approach can still produce a correct solution 5.9\% of the time, whereas all baseline methods fail.}
These low $N_c$ scenarios serve as a proxy for a class of most challenging problems where generating a correct solution is difficult. The effectiveness of our method in these situations highlights its robustness for solving difficult problems.

\paragraph{Generalization across Benchmarks} As shown in Figure~\ref{fig:correct_ratio}, GSR-7B is able to recover a correct solution from a complete set of incorrect candidates ($N_c=0$) across all benchmarks, ranging from 3.4\% on MATH500 to 9.0\% on AMC22-23. This consistency proves the refinement capability is a generalizable skill that GSR-7B has acquired.

\subsection{Robustness and Generalization Analysis}

In this section, we conduct analytic experiments on two larger models, Qwen2.5-14B/32B-Instruct  \citep{qwen2.5} to address two critical questions: (\textbf{RQ3}) \textit{Is our method effective across different model scales?} (\textbf{RQ4}) \textit{Does the model learn a generalizable improvement skill, or does it merely learn to correct its own specific errors?} We use the same hybrid training dataset and experimental setup as described in Section \ref{sec:ablation dual-training}. 

\begin{table}[ht]
\centering
\caption{Performance on AIME benchmarks across different model scales. The table compares two fine-tuned models (w training) against their base counterparts (w/o training).}
\label{tab:model_size}
\resizebox{\linewidth}{!}{%

\begin{tabular}{lcccc}
\toprule
\multirow{2}{*}{\textbf{Method}} 
  & \multicolumn{2}{c}{\makecell{\textbf{Qwen2.5-}\\\textbf{14B-Instruct}}}
  & \multicolumn{2}{c}{\makecell{\textbf{Qwen2.5-}\\\textbf{32B-Instruct}}} \\
\cmidrule(lr){2-3} \cmidrule(lr){4-5}
&\textbf{AIME24} & \textbf{AIME25} & \textbf{AIME24} & \textbf{AIME25} \\
    \midrule
    \multicolumn{5}{c}{w/o Training} \\
    \midrule

pass@1    & 13.5          & 12.7          & 16.7          & 12.9 \\
maj@4     & \textbf{17.5} & \textbf{15.8} & 19.2          & 15.8 \\
selfRef@4 & 16.0          & \textbf{15.8} & \textbf{22.3} & \textbf{16.3} \\
    \midrule
    \multicolumn{5}{c}{w Training} \\
    \midrule
pass@1    & 49.4          & 36.9          & 66.3          & 54.2 \\
maj@4     & 56.7          & 41.7          & 72.5          & 59.2 \\
\rowcolor{blue!6} selfRef@4 & \textbf{68.1} & \textbf{49.6} & \textbf{75.2} & \textbf{67.3} \\
\bottomrule
\end{tabular}
}%
\end{table}

\paragraph{Robustness Across Different Model Sizes}
The results, presented in Table \ref{tab:model_size}, demonstrate the effective scalability of our approach to models of a larger-scale.
After training, both the 14B and 32B models exhibit substantial performance improvements across standard metrics (pass@1 and maj@4) compared to their w/o SFT counterparts.
More critically, the results demonstrate the effectiveness of our method. For fine-tuned models, selfRef@4 consistently outperforms their respective majority voting baselines. For instance, the fine-tuned 14B model achieves a 20.1\% relative improvement (68.1\% vs. 56.7\%) in AIME24, while the 32B model shows a 13.7\% relative improvement (67.3\% vs. 59.2\%) in AIME25. This phenomenon contrasts with the 14B base model, which fails to benefit from self-refinement and even shows a performance degradation in AIME24 compared to the majority voting baseline. 

\paragraph{Generalization via Data Decoupling}
A crucial aspect of this experiment is that we reuse the hybrid dataset from Section \ref{sec:ablation dual-training} to train larger models. In particular, \textbf{all candidate solutions within this dataset are generated by Qwen2.5-7B-Instruct}. Although there is a mismatch between the candidate generator and learners, the results clearly indicate that both models successfully acquire the ability of self-refinement. 
This finding provides compelling evidence for a decoupling of the model's internal knowledge representation from the self-refinement skill. 

\subsection{Input Scaling Analysis}

To scale up to a larger number of inputs, previous work \citep{zhang2025cotbasedsynthesizerenhancingllm, guo2025rewardreasoningmodel} has often employed a hierarchical strategy, partitioning candidate responses into fixed-size groups, producing outputs for each group, and then iteratively combining across groups. This approach has been shown to be general and effective. However, such methods are not indeed scaling on input and have a time complexity of $\mathcal{O}(N)$.

\begin{figure}[ht]
\centering
\includegraphics[width=1.0\columnwidth]{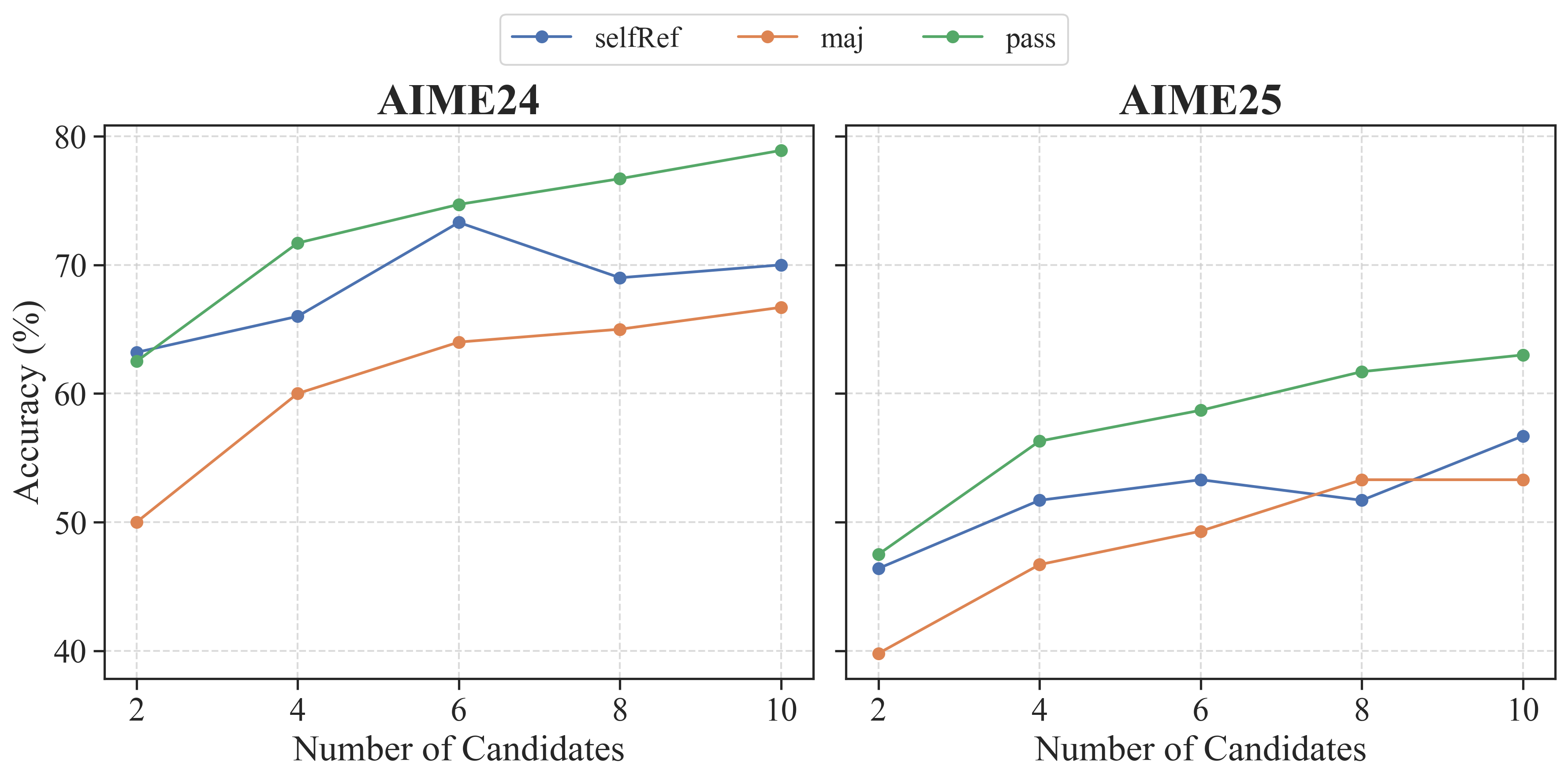} 
\caption{Accuracy of input scaling on the performance across the (\textbf{\textit{Left}}) AIME24 and (\textbf{\textit{Right}}) AIME25 benchmarks.
}
\label{fig:input scaling}
\end{figure}

Our experiments are conducted on \textbf{a fixed set of four candidate responses}. To evaluate our model's extrapolation capability, we conduct an input scaling experiment. 
Specifically, we evaluate performance with the number of candidate responses, $k$, ranging from 2 to 10.
We compare selfRef@k (self-refinement) with the maj@k (majority voting) and oracle pass@k. 
In Figure \ref{fig:input scaling}, for a smaller number of candidates ($k\leq4$), GSR-7B shows strong performance, even slightly \textbf{surpassing the pass@2} (63.2\% vs. 62.5\%) on AIME24 at $k=2$.
For a larger number of candidates, our method consistently outperforms majority voting, and its performance scales with the number of candidates. 
This suggests that our method effectively extrapolates to a number of candidates greater than four. 

\subsection{Out-of-Distribution Analysis}

\begin{figure}[ht]
\centering
\includegraphics[width=1.0\columnwidth]{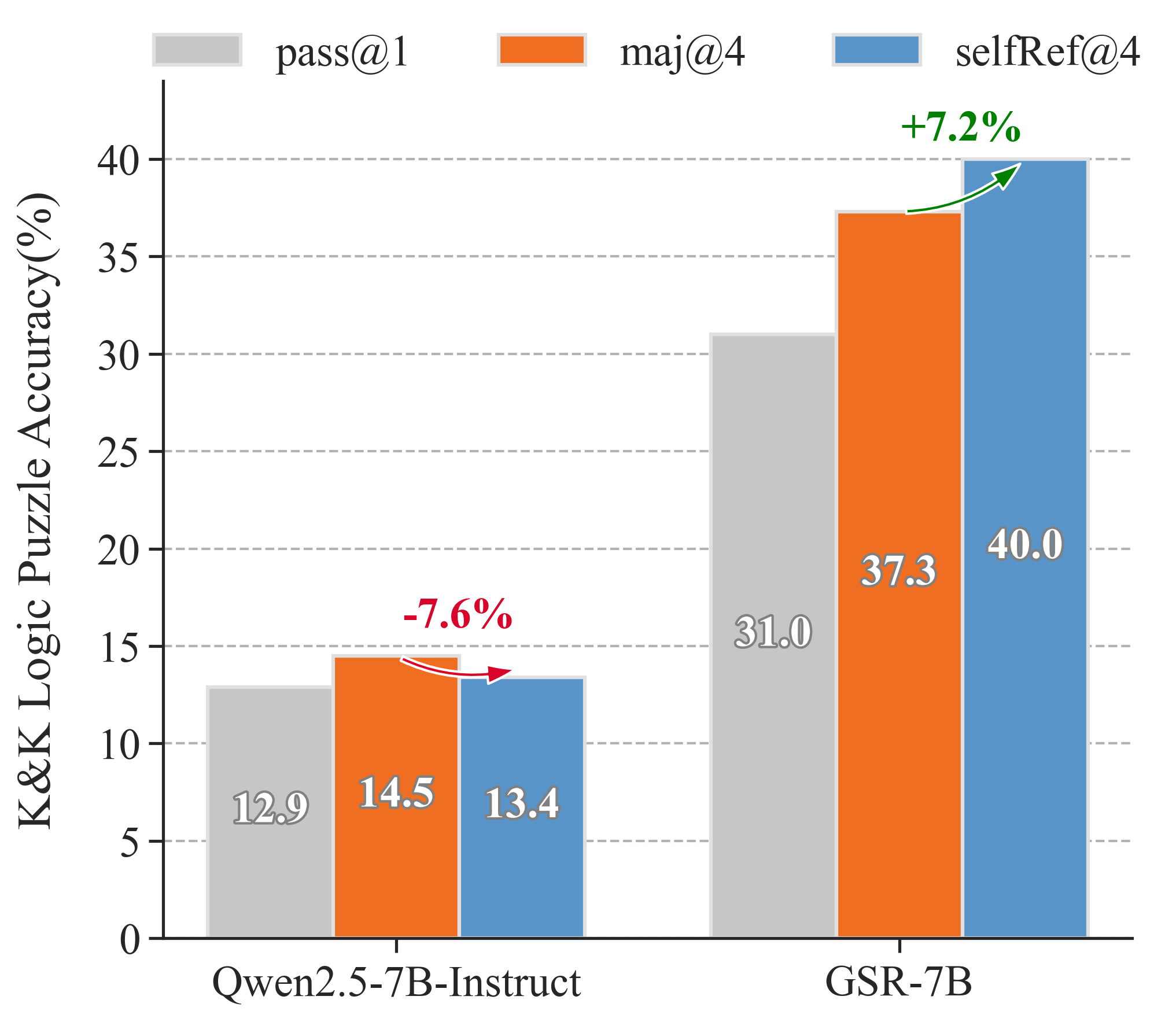} 
\caption{
Experiments results for our model and its base model on a subset of K\&K dataset. The arrows and percentages quantify the relative performance change when applying our method over majority voting.
}
\label{fig:kk_fig}
\end{figure}


GSR-7B is trained exclusively on math datasets. To further analyze its generalization ability to an unseen domain, we test it on the Knights and Knaves (K\&K) logic puzzles dataset \citep{xie2025memorizationlargelanguagemodels}, where some characters tell the truth and others only lie. 
For evaluation, we use the 4ppl subset of the K\&K dataset and reformat original problems as multiple-choice questions to allow more accurate scoring.
We report the performance of pass@1, majority voting (maj@4) and our method (selfRef@4).
In Figure \ref{fig:kk_fig}, GSR-7B demonstrates a substantial performance improvement over the base model. Compared to the base model's pass@1 accuracy of 12.9\%, our model achieves 37.3\%. More critically, we observe a clear divergence in self-refinement ability. 
The base model's selfRef@4 performance decreases by 7.6\% relative to maj@4, while our model shows a performance improvement of 7.2\%.
These results confirm that the self-refinement ability is a skill acquired during training and that this skill even successfully generalizes to an unseen out-of-distribution domain.

\section{Related Work}

\paragraph{Aggregation Methods}
The most well-known approach within parallel aggregation is majority voting \citep{wang2023selfconsistency}, which leverages the self-consistency of the model's outputs to select the most consistent answer.
Best-of-N (BoN) replaces the voting mechanism with an external verifier (Reward Model, RM), which is used to either assign a scalar score \citep{cobbe2021trainingverifierssolvemath, gao2022scalinglawsrewardmodel, MetaMetrics}, or to perform discriminative methods to indicate a preference \citep{Stiennon2020summarizewithhumanfeedback, nakano2022webgptbrowserassistedquestionansweringhuman}. 
Some works have introduced the Generative Reward Model (GenRM), which leverages the model's generative capacity to provide a detailed and interpretable justification \citep{Kim2024Prometheus2, mahan2024generativerewardmodels, whitehouse2025j1incentivizingthinkingllmasajudge, chen2025rmr1rewardmodelingreasoning, guo2025rewardreasoningmodel}.

\paragraph{Refinement Methods}

Some studies have explored that models can refine responses to improve the performance \citep{cobbe2021trainingverifierssolvemath, gou2024criticlargelanguagemodels, ferraz2024llmselfcorrectiondecrimdecompose}.
Self-Refine \citep{madaan2023selfrefineiterativerefinementselffeedback}, RIC \citep{kim2023languagemodelssolvecomputer} and REFINER \citep{paul2024refinerreasoningfeedbackintermediate} focus on sequential self-refinement, where a single model is used to generate feedback on its own output and iteratively revise the solution accordingly.
Distinct from sequential approaches, LLM-Blender \citep{jiang-etal-2023-llm}, LMCor \citep{vernikos-etal-2024-small}, AFT \citep{li2025draftsanswersunlockingllm}, CoT-based-Synthesizer \citep{zhang2025cotbasedsynthesizerenhancingllm} train a dedicated generative model to fuse multiple solutions. However, these methods focus exclusively on the task of improving parallel solutions, while neglecting to train the model to solve problems directly. 
Besides, \citet{zhao2025majorityrightrltraining} learns solution aggregation as an explicit skill via reinforcement learning, showing that majority voting is not always optimal.
MoA \citep{wang2024mixtureofagentsenhanceslargelanguage} and Multiagent FT \citep{subramaniam2025multiagentfinetuningselfimprovement} leverages multi-agent society of specialized LLMs to improve responses, incurring significant inference and deployment overhead.

\section{Conclusion}
In this paper, we systematically investigate the scaling properties of the gains of self-refinement over majority voting in LLMs. By introducing Refinement Gap, we provide a clear scaling trend: relative refinement gap increases with model size and is only weakly related to base capability.
Motivated by this, we propose Generative Self-Refinement (GSR), a parallel TTS framework that transfers refinement ability from teacher models to smaller student models. GSR jointly optimizes for direct solving and refinement objectives, enabling a single and unified model to generate strong candidates and synthesize a better final answer even from imperfect candidates. 
Empirically, GSR substantially improves small models reasoning performance and demonstrates robust refinement behavior under challenging candidate conditions, highlighting refinement as a learnable inference strategy that can be explicitly activated via training.

\section{Limitations}
Due to limited computational resources, our experiments focus on mathematical reasoning domains. Although this setting is natural for studying self-refinement and test-time scaling, it may not fully represent other reasoning scenarios (e.g., logical deduction, programming, scientific QA, or agentic tool use). We plan to conduct more experiments across a broader range of reasoning domains in the future work.

Our data construction pipeline relies on strong teacher models and the verifiable reward to filter solutions. This design may be less directly applicable to tasks that lacks reliable verification. Extending our approach to such settings likely requires alternative supervision signals, such as human annotation, preference-based feedback, or external verifiers/tools, which may introduce additional cost and noise.


\bibliography{custom}

@misc{moshkov2025aimo,
    title={Aimo-2 winning solution: Building state-of-the-art mathematical reasoning models with openmathreasoning dataset},
    author={Moshkov, Ivan and Hanley, Darragh and Sorokin, Ivan and Toshniwal, Shubham and Henkel, Christof and Schifferer, Benedikt and Du, Wei and Gitman, Igor},
    year={2025},
    eprint={2504.16891},
    archivePrefix={arXiv},
    primaryClass={cs.AI},
    url={https://arxiv.org/abs/2504.16891}, 
    }

@misc{zhang2025cotbasedsynthesizerenhancingllm,
      title={CoT-based Synthesizer: Enhancing LLM Performance through Answer Synthesis}, 
      author={Bohan Zhang and Xiaokang Zhang and Jing Zhang and Jifan Yu and Sijia Luo and Jie Tang},
      year={2025},
      eprint={2501.01668},
      archivePrefix={arXiv},
      primaryClass={cs.CL},
      url={https://arxiv.org/abs/2501.01668}, 
}

@inproceedings{wang2023selfconsistency,
  author       = {Xuezhi Wang and
                  Jason Wei and
                  Dale Schuurmans and
                  Quoc V. Le and
                  Ed H. Chi and
                  Sharan Narang and
                  Aakanksha Chowdhery and
                  Denny Zhou},
  title        = {Self-Consistency Improves Chain of Thought Reasoning in Language Models},
  booktitle    = {Proceedings of {ICLR} 2023},
  year         = {2023}
}

@misc{liu2024skyworkrewardbagtricksreward,
      title={Skywork-Reward: Bag of Tricks for Reward Modeling in LLMs}, 
      author={Chris Yuhao Liu and Liang Zeng and Jiacai Liu and Rui Yan and Jujie He and Chaojie Wang and Shuicheng Yan and Yang Liu and Yahui Zhou},
      year={2024},
      eprint={2410.18451},
      archivePrefix={arXiv},
      primaryClass={cs.AI},
      url={https://arxiv.org/abs/2410.18451}, 
}

@misc{guo2025rewardreasoningmodel,
      title={Reward Reasoning Model}, 
      author={Jiaxin Guo and Zewen Chi and Li Dong and Qingxiu Dong and Xun Wu and Shaohan Huang and Furu Wei},
      year={2025},
      eprint={2505.14674},
      archivePrefix={arXiv},
      primaryClass={cs.CL},
      url={https://arxiv.org/abs/2505.14674}, 
}

@misc{chen2025rmr1rewardmodelingreasoning,
      title={RM-R1: Reward Modeling as Reasoning}, 
      author={Xiusi Chen and Gaotang Li and Ziqi Wang and Bowen Jin and Cheng Qian and Yu Wang and Hongru Wang and Yu Zhang and Denghui Zhang and Tong Zhang and Hanghang Tong and Heng Ji},
      year={2025},
      eprint={2505.02387},
      archivePrefix={arXiv},
      primaryClass={cs.CL},
      url={https://arxiv.org/abs/2505.02387}, 
}

@inproceedings{hendrycks2021measuringmathematicalproblemsolving,
  author       = {Dan Hendrycks and
                  Collin Burns and
                  Saurav Kadavath and
                  Akul Arora and
                  Steven Basart and
                  Eric Tang and
                  Dawn Song and
                  Jacob Steinhardt},
  title        = {Measuring Mathematical Problem Solving With the {MATH} Dataset},
  booktitle    = {Proceedings of {NeurIPS} 2021},
  year         = {2021}
}

@inproceedings{he2024olympiadbenchchallengingbenchmarkpromoting,
  author       = {Chaoqun He and
                  Renjie Luo and
                  Yuzhuo Bai and
                  Shengding Hu and
                  Zhen Leng Thai and
                  Junhao Shen and
                  Jinyi Hu and
                  Xu Han and
                  Yujie Huang and
                  Yuxiang Zhang and
                  Jie Liu and
                  Lei Qi and
                  Zhiyuan Liu and
                  Maosong Sun},
  title        = {OlympiadBench: {A} Challenging Benchmark for Promoting {AGI} with
                  Olympiad-Level Bilingual Multimodal Scientific Problems},
  booktitle    = {Proceedings of {ACL} 2024},
  pages        = {3828--3850},
  publisher    = {Association for Computational Linguistics},
  year         = {2024}
}

@misc{aimo_validation_aime,
  author       = {AI-MO},
  title        = {{AIMO Validation AIME Dataset}},
  year         = {2024},
  publisher    = {Hugging Face},
  journal      = {Hugging Face Datasets},
  howpublished = {\url{https://huggingface.co/datasets/AI-MO/aimo-validation-aime}},
  note         = {Accessed: 2025-03-29}
}

@misc{aime2025,
  author       = {Yen-Ting Lin},
  title        = {AIME 2025 Dataset},
  year         = {2025},
  publisher    = {Hugging Face},
  journal      = {Hugging Face Datasets},
  howpublished = {\url{https://huggingface.co/datasets/yentinglin/aime_2025}},
  note         = {Accessed: 2025-03-29}
}

@misc{aimo_validation_amc,
  author       = {AI-MO},
  title        = {{AIMO Validation AMC Dataset}},
  year         = {2024},
  publisher    = {Hugging Face},
  journal      = {Hugging Face Datasets},
  howpublished = {\url{https://huggingface.co/datasets/AI-MO/aimo-validation-amc}},
  note         = {Accessed: 2025-03-29}
}

@misc{xie2025memorizationlargelanguagemodels,
      title={On Memorization of Large Language Models in Logical Reasoning}, 
      author={Chulin Xie and Yangsibo Huang and Chiyuan Zhang and Da Yu and Xinyun Chen and Bill Yuchen Lin and Bo Li and Badih Ghazi and Ravi Kumar},
      year={2025},
      eprint={2410.23123},
      archivePrefix={arXiv},
      primaryClass={cs.CL},
      url={https://arxiv.org/abs/2410.23123}, 
}

@misc{qwen2.5,
      title={Qwen2.5 Technical Report}, 
      author={An Yang and Baosong Yang and Beichen Zhang and Binyuan Hui and Bo Zheng and Bowen Yu and Chengyuan Li and Dayiheng Liu and Fei Huang and Haoran Wei and Huan Lin and Jian Yang and Jianhong Tu and Jianwei Zhang and Jianxin Yang and Jiaxi Yang and Jingren Zhou and Junyang Lin and Kai Dang and Keming Lu and Keqin Bao and Kexin Yang and Le Yu and Mei Li and Mingfeng Xue and Pei Zhang and Qin Zhu and Rui Men and Runji Lin and Tianhao Li and Tianyi Tang and Tingyu Xia and Xingzhang Ren and Xuancheng Ren and Yang Fan and Yang Su and Yichang Zhang and Yu Wan and Yuqiong Liu and Zeyu Cui and Zhenru Zhang and Zihan Qiu},
      year={2025},
      eprint={2412.15115},
      archivePrefix={arXiv},
      primaryClass={cs.CL},
      url={https://arxiv.org/abs/2412.15115}, 
}

@misc{qwen3,
      title={Qwen3 Technical Report}, 
      author={An Yang and Anfeng Li and Baosong Yang and Beichen Zhang and Binyuan Hui and Bo Zheng and Bowen Yu and Chang Gao and Chengen Huang and Chenxu Lv and Chujie Zheng and Dayiheng Liu and Fan Zhou and Fei Huang and Feng Hu and Hao Ge and Haoran Wei and Huan Lin and Jialong Tang and Jian Yang and Jianhong Tu and Jianwei Zhang and Jianxin Yang and Jiaxi Yang and Jing Zhou and Jingren Zhou and Junyang Lin and Kai Dang and Keqin Bao and Kexin Yang and Le Yu and Lianghao Deng and Mei Li and Mingfeng Xue and Mingze Li and Pei Zhang and Peng Wang and Qin Zhu and Rui Men and Ruize Gao and Shixuan Liu and Shuang Luo and Tianhao Li and Tianyi Tang and Wenbiao Yin and Xingzhang Ren and Xinyu Wang and Xinyu Zhang and Xuancheng Ren and Yang Fan and Yang Su and Yichang Zhang and Yinger Zhang and Yu Wan and Yuqiong Liu and Zekun Wang and Zeyu Cui and Zhenru Zhang and Zhipeng Zhou and Zihan Qiu},
      year={2025},
      eprint={2505.09388},
      archivePrefix={arXiv},
      primaryClass={cs.CL},
      url={https://arxiv.org/abs/2505.09388}, 
}

@misc{qwq32b,
    title = {QwQ-32B: Embracing the Power of Reinforcement Learning},
    url = {https://qwenlm.github.io/blog/qwq-32b/},
    author = {QwenTeam},
    month = {March},
    year = {2025}
}

@misc{deepseekr1,
      title={DeepSeek-R1: Incentivizing Reasoning Capability in LLMs via Reinforcement Learning}, 
      author={DeepSeek-AI},
      year={2025},
      eprint={2501.12948},
      archivePrefix={arXiv},
      primaryClass={cs.CL},
      url={https://arxiv.org/abs/2501.12948}, 
}

@misc{hochlehnert2025soberlookprogresslanguage,
      title={A Sober Look at Progress in Language Model Reasoning: Pitfalls and Paths to Reproducibility}, 
      author={Andreas Hochlehnert and Hardik Bhatnagar and Vishaal Udandarao and Samuel Albanie and Ameya Prabhu and Matthias Bethge},
      year={2025},
      eprint={2504.07086},
      archivePrefix={arXiv},
      primaryClass={cs.LG},
      url={https://arxiv.org/abs/2504.07086}, 
}

@inproceedings{jiang-etal-2023-llm,
  author       = {Dongfu Jiang and
                  Xiang Ren and
                  Bill Yuchen Lin},
  title        = {LLM-Blender: Ensembling Large Language Models with Pairwise Ranking
                  and Generative Fusion},
  booktitle    = {Proceedings of {ACL} 2023},
  pages        = {14165--14178},
  publisher    = {Association for Computational Linguistics},
  year         = {2023},
}

@inproceedings{vernikos-etal-2024-small,
  author       = {Giorgos Vernikos and
                  Arthur Brazinskas and
                  Jakub Ad{\'{a}}mek and
                  Jonathan Mallinson and
                  Aliaksei Severyn and
                  Eric Malmi},
  title        = {Small Language Models Improve Giants by Rewriting Their Outputs},
  booktitle    = {Proceedings of {EACL} 2024},
  pages        = {2703--2718},
  publisher    = {Association for Computational Linguistics},
  year         = {2024},
}

@misc{brown2024largelanguagemonkeysscaling,
      title={Large Language Monkeys: Scaling Inference Compute with Repeated Sampling}, 
      author={Bradley Brown and Jordan Juravsky and Ryan Ehrlich and Ronald Clark and Quoc V. Le and Christopher Ré and Azalia Mirhoseini},
      year={2024},
      eprint={2407.21787},
      archivePrefix={arXiv},
      primaryClass={cs.LG},
      url={https://arxiv.org/abs/2407.21787}, 
}

@misc{snell2024scalingllmtesttimecompute,
      title={Scaling LLM Test-Time Compute Optimally can be More Effective than Scaling Model Parameters}, 
      author={Charlie Snell and Jaehoon Lee and Kelvin Xu and Aviral Kumar},
      year={2024},
      eprint={2408.03314},
      archivePrefix={arXiv},
      primaryClass={cs.LG},
      url={https://arxiv.org/abs/2408.03314}, 
}

@misc{irvine2023rewardingchatbotsrealworldengagement,
      title={Rewarding Chatbots for Real-World Engagement with Millions of Users}, 
      author={Robert Irvine and Douglas Boubert and Vyas Raina and Adian Liusie and Ziyi Zhu and Vineet Mudupalli and Aliaksei Korshuk and Zongyi Liu and Fritz Cremer and Valentin Assassi and Christie-Carol Beauchamp and Xiaoding Lu and Thomas Rialan and William Beauchamp},
      year={2023},
      eprint={2303.06135},
      archivePrefix={arXiv},
      primaryClass={cs.CL},
      url={https://arxiv.org/abs/2303.06135}, 
}

@inproceedings{song2024goodbadgreedyevaluation,
  author       = {Yifan Song and
                  Guoyin Wang and
                  Sujian Li and
                  Bill Yuchen Lin},
  title        = {The Good, The Bad, and The Greedy: Evaluation of LLMs Should Not Ignore
                  Non-Determinism},
  booktitle    = {Proceedings of {NAACL} 2025},
  pages        = {4195--4206},
  publisher    = {Association for Computational Linguistics},
  year         = {2025},
}

@misc{cobbe2021trainingverifierssolvemath,
      title={Training Verifiers to Solve Math Word Problems}, 
      author={Karl Cobbe and Vineet Kosaraju and Mohammad Bavarian and Mark Chen and Heewoo Jun and Lukasz Kaiser and Matthias Plappert and Jerry Tworek and Jacob Hilton and Reiichiro Nakano and Christopher Hesse and John Schulman},
      year={2021},
      eprint={2110.14168},
      archivePrefix={arXiv},
      primaryClass={cs.LG},
      url={https://arxiv.org/abs/2110.14168}, 
}

@inproceedings{gao2022scalinglawsrewardmodel,
  author       = {Rafael Rafailov and
                  Yaswanth Chittepu and
                  Ryan Park and
                  Harshit Sikchi and
                  Joey Hejna and
                  W. Bradley Knox and
                  Chelsea Finn and
                  Scott Niekum},
  title        = {Scaling Laws for Reward Model Overoptimization in Direct Alignment
                  Algorithms},
  booktitle    = {Proceedings of {NeurIPS} 2024},
  year         = {2024}
}

@inproceedings{MetaMetrics,
  author       = {Genta Indra Winata and
                  David Anugraha and
                  Lucky Susanto and
                  Garry Kuwanto and
                  Derry Tanti Wijaya},
  title        = {MetaMetrics: Calibrating Metrics for Generation Tasks Using Human
                  Preferences},
  booktitle    = {Proceedings of {ICLR} 2025},
  year         = {2025},
}

@inproceedings{Stiennon2020summarizewithhumanfeedback,
  author       = {Nisan Stiennon and
                  Long Ouyang and
                  Jeffrey Wu and
                  Daniel M. Ziegler and
                  Ryan Lowe and
                  Chelsea Voss and
                  Alec Radford and
                  Dario Amodei and
                  Paul F. Christiano},
  title        = {Learning to summarize with human feedback},
  booktitle    = {Proceedings of {NeurIPS} 2020},
  year         = {2020}
}

@misc{nakano2022webgptbrowserassistedquestionansweringhuman,
      title={WebGPT: Browser-assisted question-answering with human feedback}, 
      author={Reiichiro Nakano and Jacob Hilton and Suchir Balaji and Jeff Wu and Long Ouyang and Christina Kim and Christopher Hesse and Shantanu Jain and Vineet Kosaraju and William Saunders and Xu Jiang and Karl Cobbe and Tyna Eloundou and Gretchen Krueger and Kevin Button and Matthew Knight and Benjamin Chess and John Schulman},
      year={2022},
      eprint={2112.09332},
      archivePrefix={arXiv},
      primaryClass={cs.CL},
      url={https://arxiv.org/abs/2112.09332}, 
}

@inproceedings{Kim2024Prometheus2,
  author       = {Seungone Kim and
                  Juyoung Suk and
                  Shayne Longpre and
                  Bill Yuchen Lin and
                  Jamin Shin and
                  Sean Welleck and
                  Graham Neubig and
                  Moontae Lee and
                  Kyungjae Lee and
                  Minjoon Seo},
  title        = {Prometheus 2: An Open Source Language Model Specialized in Evaluating
                  Other Language Models},
  booktitle    = {Proceedings of {EMNLP} 2024},
  pages        = {4334--4353},
  publisher    = {Association for Computational Linguistics},
  year         = {2024},
}

@misc{mahan2024generativerewardmodels,
      title={Generative Reward Models}, 
      author={Dakota Mahan and Duy Van Phung and Rafael Rafailov and Chase Blagden and Nathan Lile and Louis Castricato and Jan-Philipp Fränken and Chelsea Finn and Alon Albalak},
      year={2024},
      eprint={2410.12832},
      archivePrefix={arXiv},
      primaryClass={cs.LG},
      url={https://arxiv.org/abs/2410.12832}, 
}

@misc{whitehouse2025j1incentivizingthinkingllmasajudge,
      title={J1: Incentivizing Thinking in LLM-as-a-Judge via Reinforcement Learning}, 
      author={Chenxi Whitehouse and Tianlu Wang and Ping Yu and Xian Li and Jason Weston and Ilia Kulikov and Swarnadeep Saha},
      year={2025},
      eprint={2505.10320},
      archivePrefix={arXiv},
      primaryClass={cs.CL},
      url={https://arxiv.org/abs/2505.10320}, 
}

@misc{yang2025reasonfluxhierarchicalllmreasoning,
      title={ReasonFlux: Hierarchical LLM Reasoning via Scaling Thought Templates}, 
      author={Ling Yang and Zhaochen Yu and Bin Cui and Mengdi Wang},
      year={2025},
      eprint={2502.06772},
      archivePrefix={arXiv},
      primaryClass={cs.CL},
      url={https://arxiv.org/abs/2502.06772}, 
}

@inproceedings{madaan2023selfrefineiterativerefinementselffeedback,
  author       = {Aman Madaan and
                  Niket Tandon and
                  Prakhar Gupta and
                  Skyler Hallinan and
                  Luyu Gao and
                  Sarah Wiegreffe and
                  Uri Alon and
                  Nouha Dziri and
                  Shrimai Prabhumoye and
                  Yiming Yang and
                  Shashank Gupta and
                  Bodhisattwa Prasad Majumder and
                  Katherine Hermann and
                  Sean Welleck and
                  Amir Yazdanbakhsh and
                  Peter Clark},
  title        = {Self-Refine: Iterative Refinement with Self-Feedback},
  booktitle    = {Proceedings of {NeurIPS} 2023},
  year         = {2023},
}

@inproceedings{kim2023languagemodelssolvecomputer,
  author       = {Geunwoo Kim and
                  Pierre Baldi and
                  Stephen McAleer},
  title        = {Language Models can Solve Computer Tasks},
  booktitle    = {Proceedings of {NeurIPS} 2023},
  year         = {2023},
}

@inproceedings{paul2024refinerreasoningfeedbackintermediate,
  author       = {Debjit Paul and
                  Mete Ismayilzada and
                  Maxime Peyrard and
                  Beatriz Borges and
                  Antoine Bosselut and
                  Robert West and
                  Boi Faltings},
  title        = {{REFINER:} Reasoning Feedback on Intermediate Representations},
  booktitle    = {Proceedings of {EACL} 2024},
  pages        = {1100--1126},
  publisher    = {Association for Computational Linguistics},
  year         = {2024},
}

@inproceedings{wang2024mixtureofagentsenhanceslargelanguage,
  author       = {Junlin Wang and
                  Jue Wang and
                  Ben Athiwaratkun and
                  Ce Zhang and
                  James Zou},
  title        = {Mixture-of-Agents Enhances Large Language Model Capabilities},
  booktitle    = {Proceedings of {ICLR} 2025},
  year         = {2025},
}

@misc{li2025draftsanswersunlockingllm,
      title={From Drafts to Answers: Unlocking LLM Potential via Aggregation Fine-Tuning}, 
      author={Yafu Li and Zhilin Wang and Tingchen Fu and Ganqu Cui and Sen Yang and Yu Cheng},
      year={2025},
      eprint={2501.11877},
      archivePrefix={arXiv},
      primaryClass={cs.CL},
      url={https://arxiv.org/abs/2501.11877}, 
}

@inproceedings{subramaniam2025multiagentfinetuningselfimprovement,
  author       = {Vighnesh Subramaniam and
                  Yilun Du and
                  Joshua B. Tenenbaum and
                  Antonio Torralba and
                  Shuang Li and
                  Igor Mordatch},
  title        = {Multiagent Finetuning: Self Improvement with Diverse Reasoning Chains},
  booktitle    = {Proceedings of {ICLR} 2025},
  year         = {2025},
}

@inproceedings{huang2024largelanguagemodelsselfcorrect,
  author       = {Jie Huang and
                  Xinyun Chen and
                  Swaroop Mishra and
                  Huaixiu Steven Zheng and
                  Adams Wei Yu and
                  Xinying Song and
                  Denny Zhou},
  title        = {Large Language Models Cannot Self-Correct Reasoning Yet},
  booktitle    = {Proceedings of {ICLR} 2024},
  year         = {2024},
}

@inproceedings{gou2024criticlargelanguagemodels,
  author       = {Zhibin Gou and
                  Zhihong Shao and
                  Yeyun Gong and
                  Yelong Shen and
                  Yujiu Yang and
                  Nan Duan and
                  Weizhu Chen},
  title        = {{CRITIC:} Large Language Models Can Self-Correct with Tool-Interactive
                  Critiquing},
  booktitle    = {Proceedings of {ICLR} 2024},
  year         = {2024},
}

@inproceedings{ferraz2024llmselfcorrectiondecrimdecompose,
  author       = {Thomas Palmeira Ferraz and
                  Kartik Mehta and
                  Yu{-}Hsiang Lin and
                  Haw{-}Shiuan Chang and
                  Shereen Oraby and
                  Sijia Liu and
                  Vivek Subramanian and
                  Tagyoung Chung and
                  Mohit Bansal and
                  Nanyun Peng},
  editor       = {Yaser Al{-}Onaizan and
                  Mohit Bansal and
                  Yun{-}Nung Chen},
  title        = {{LLM} Self-Correction with DeCRIM: Decompose, Critique, and Refine
                  for Enhanced Following of Instructions with Multiple Constraints},
  booktitle    = {Findings of Proceedings of {EMNLP} 2024},
  pages        = {7773--7812},
  publisher    = {Association for Computational Linguistics},
  year         = {2024},
}

@misc{zhao2025majorityrightrltraining,
      title={The Majority is not always right: RL training for solution aggregation}, 
      author={Wenting Zhao and Pranjal Aggarwal and Swarnadeep Saha and Asli Celikyilmaz and Jason Weston and Ilia Kulikov},
      year={2025},
      eprint={2509.06870},
      archivePrefix={arXiv},
      primaryClass={cs.CL},
      url={https://arxiv.org/abs/2509.06870}, 
}

@misc{li2025llmsgeneratebetteranswer,
      title={LLMs Can Generate a Better Answer by Aggregating Their Own Responses}, 
      author={Zichong Li and Xinyu Feng and Yuheng Cai and Zixuan Zhang and Tianyi Liu and Chen Liang and Weizhu Chen and Haoyu Wang and Tuo Zhao},
      year={2025},
      eprint={2503.04104},
      archivePrefix={arXiv},
      primaryClass={cs.CL},
      url={https://arxiv.org/abs/2503.04104}, 
}

@book{kuncheva2014combining,
  title={Combining pattern classifiers: methods and algorithms},
  author={Kuncheva, Ludmila I},
  year={2014},
  publisher={John Wiley \& Sons}
}

@misc{khairi2025makingtakingbestn,
      title={Making, not Taking, the Best of N}, 
      author={Ammar Khairi and Daniel D'souza and Marzieh Fadaee and Julia Kreutzer},
      year={2025},
      eprint={2510.00931},
      archivePrefix={arXiv},
      primaryClass={cs.CL},
      url={https://arxiv.org/abs/2510.00931}, 
}

\appendix
\clearpage
\section{Prompt Template}
\label{sec:prompt}

GSR supports dual abilities during inference: direct-answer generation and self-refinement. The choice between modes is achieved by constructing different input prompts. Here we provide the detailed prompt template used to prompt the model to perform self-refinement. The prompt template consists of the original problem and a set of candidate solutions mainly.

\begin{tcolorbox}[
breakable,
  title=Prompt Template on the Refinement Mode,
  colback=blue!5,
  colframe=blue!50!white,
  fonttitle=\bfseries,
  coltitle=white,
  colbacktitle=gray!70!blue,
  boxrule=0.5pt,
  arc=2mm,
]
You are an expert and creative solver, provided with a challenging problem and a set of candidate responses which may be correct, partially correct or even wrong.

You should first fully summarize the connection between candidate responses and problem, then generate a new and superior solution. You should generate a correct solution yourself if all candidates are wrong. Don't copy candidates, use insights selectively and reason independently. 

Problem:    \\
\{Problem\} \\
Think step by step and put final answer within \texttt{\textbackslash boxed\{\}}.\\

Candidate Response 1:   \\
\{Response 1\}  \\

Candidate Response 2:   \\
\{Response 2\}  \\

Candidate Response 3:   \\
\{Response 3\}  \\

Candidate Response 4:   \\
\{Response 4\}  
\end{tcolorbox}

We also provide a common prompt template to instruct the model to solve the problem directly. This common prompt template is applicable for any LLMs.

\begin{tcolorbox}[
  title=Common Prompt Template for Direct-Answer Generation,
  colback=blue!5,
  colframe=blue!50!white,
  fonttitle=\bfseries,
  coltitle=white,
  colbacktitle=gray!70!blue,
  boxrule=0.5pt,
  arc=2mm,
]
\{Problem\} \\
Think step by step and put final answer within \texttt{\textbackslash boxed\{\}}.
\end{tcolorbox}

\section{Analysis of Candidate Input Burden}
\label{sec:analysis_input_burden}
\begin{figure}[ht]
\centering
\includegraphics[width=1.0\columnwidth]{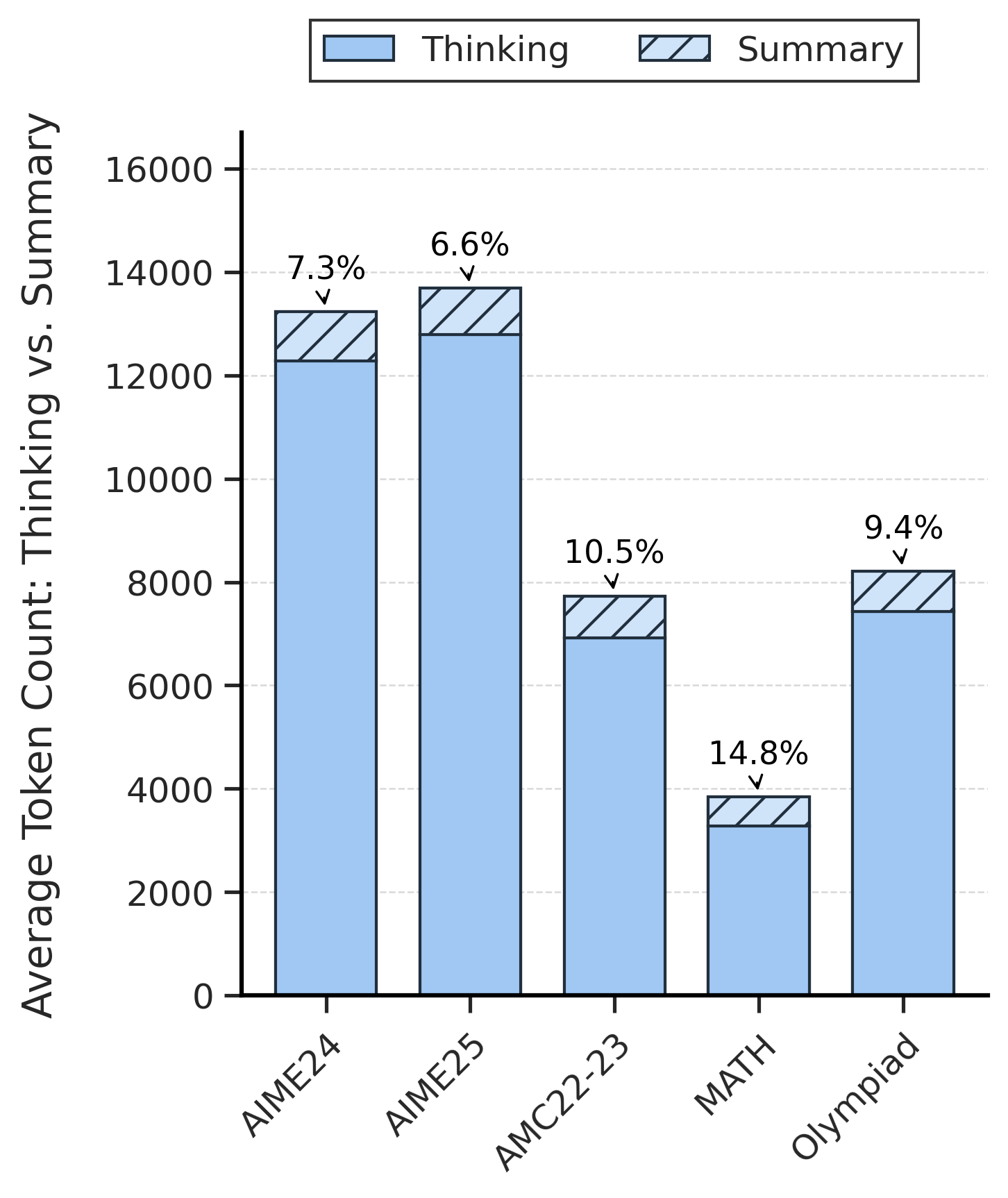} 
\caption{
Average token counts of model \textbf{Direct-Solving} outputs across five benchmarks, showing the breakdown between the Thinking (solid) and Summary (hatched) components. Percentages indicate the proportion of summary tokens.
}
\label{fig:token count}
\end{figure}

\begin{table*}[ht!]
\centering
\caption{Average token counts of model \textbf{Self-Refinement} outputs against \textbf{Direct Solving} across five benchmarks.}
\label{tab:selfR_token_count}
\begin{tabular}{llcccc}
\toprule
\diagbox[font=\bfseries]{Token Count}{Benchmark}
& \textbf{AIME24} & \textbf{AIME25} & \textbf{AMC22-23} & \textbf{MATH500} & \textbf{Olympiad} \\

\midrule
\textbf{self-Refinement} & 8878 & 10477 & 5270 & 2745 & 5910 \\
\textbf{Direct Solving} & 13232 & 13687 & 7732 & 3850 & 8209 \\
\textbf{Reduction(\%)} & -32.9\% & -23.5\% & -31.8\% & -28.7\% & -28.0\% \\ 
\bottomrule
\end{tabular}
\end{table*}

A critical factor for the feasibility of our method is the management of context length, as recontextualizing the verbose responses from thinking model can lead to prohibitive compute overhead. We innovatively mitigate this by exclusively \textbf{extracting the summary component from responses while discarding the preceding thinking component}.

To quantify the efficiency of this strategy, we analyze the token composition of our model's outputs across the five benchmarks of Section \ref{sec:main results}. As illustrated in Figure \ref{fig:token count}, the summary component is remarkably succinct, consistently accounting for only a small fraction of the total generated tokens (e.g., 6.6\% to 14.8\%). In particular, for difficult benchmarks like AIME24 and AIME25, the total number of tokens generated is substantial and exceeds 13,000 on average. Despite this, the summary component itself still remains remarkably concise, with an average token count of fewer than 1,000. Across the five distinct and representative benchmarks, the token length of the summary component varies from 570 to 960 tokens. 
This demonstrates that the input context is well controlled, imposing an average input burden of less than 4000 tokens even on the challenging AIME24 benchmark.

As detailed in Table \ref{tab:selfR_token_count}, we further calculate the average token length for GSR-7B to perform self-refinement given four candidate responses. We find that token consumption is significantly lower than the direct-solving approach. We attribute this efficiency to the model's ability to leverage valuable information from correct candidate responses and thus to prune the search space significantly during its chain-of-thought reasoning.



\section{Full Results of Fine-Grained Analysis }
\label{sec:fine_grained_full_results}
In this section, we provide the full experimental results in Section \ref{sec:fine-grained analysis}. 
The results of fine-grained analysis of our method across AIME24, AIME25, AMC22-23, MATH500, Olympiad is provided in Table \ref{tab:full_correct_ratio}.
For each number of correct candidate responses (from 0 to 4), we compile statistics on our model's performance, including the number of correct answers, wrong answers, total trials, and the final correct ratio.

\begin{table*}[ht]
\centering
  \caption{  Full results of robustness analysis of our method across five mathematical benchmarks. We report correct number, wrong number, correct ratio and total conditioned on the number of correct candidates.   }
  \label{tab:full_correct_ratio}
    \resizebox{\linewidth}{!}{

    \begin{tabular}{cllllll}
    \toprule
    \textbf{Number of}  & \multirow{2}{*}{\textbf{Statistics}} & \multicolumn{5}{c}{\textbf{Benchmarks}}  \\
    \textbf{Correct Candidates} &   & \textbf{AIME24}   & \textbf{AIME25}   & \textbf{AMC22-23} & \textbf{MATH500}  & \textbf{Olympiad} \\
    \midrule
    \multirow{4}{*}{4/4}  
                        & Correct           & 260      & 200      & 867      & 6604     & 5573     \\
                        & Wrong             & 0        & 0        & 1        & 12       & 11       \\
                        & Total             & 260      & 200      & 868      & 6616     & 5584     \\
                        & Correct Ratio(\%) & 100      & 100      & 99.9     & 99.8     & 99.8 \\
    \midrule
    \multirow{4}{*}{3/4}                               
                        & Correct           & 152      & 92       & 147      & 581      & 1049     \\
                        & Wrong             & 4        & 4        & 1        & 35       & 67       \\
                        & Total             & 156      & 96       & 148      & 616      & 1116     \\
                        & Correct Ratio(\%) & 97.4     & 95.8     & 99.3     & 94.3     & 94.0 \\
    \midrule
    \multirow{4}{*}{2/4}
                        & Correct           & 129      & 101      & 80       & 165      & 564      \\
                        & Wrong             & 15       & 15       & 20       & 51       & 132      \\
                        & Total             & 144      & 116      & 100      & 216      & 696      \\
                        & Correct Ratio(\%) & 89.6     & 87.1     & 80       & 76.4     & 81.0 \\
    \midrule
    \multirow{4}{*}{1/4}
                        & Correct           & 77       & 84       & 30       & 112      & 367      \\
                        & Wrong             & 51       & 44       & 26       & 116      & 357      \\
                        & Total             & 128      & 128      & 56       & 228      & 724      \\
                        & Correct Ratio(\%) & 60.2     & 65.6     & 53.6     & 49.1     & 50.7 \\
    \midrule
    \multirow{4}{*}{0/4}  
                        & Correct           & 16       & 19       & 14       & 11       & 117      \\
                        & Wrong             & 256      & 401      & 142      & 313      & 2563     \\
                        & Total             & 272      & 420      & 156      & 324      & 2680     \\
                        & Correct Ratio(\%) & 5.9      & 4.5      & 9.0      & 3.4      & 4.4 \\
    \bottomrule
    \end{tabular}
    }
\end{table*}

\section{Experiments Details}
\label{sec:experiments details}

\subsection{Training Settings}
\label{sec:training_settings}
We train Qwen2.5-7B-Instruct\citep{qwen2.5} on the 368K SFT dataset for 3 epochs with the AdamW optimizer, employing a 10\% linear warmup followed by a cosine learning rate decay schedule. The maximum learning rate is set to 1e-4, with a batch size of 128 and a maximum sequence length of 24K tokens.
To support longer context windows and align with advanced thinking mode, we adopted the chat\_template from the Qwen3 family\citep{qwen3} and extended the maximum sequence length by setting max\_position\_embeddings to 65,536.

In the ablation study, we train the model for 5 epochs and set the maximum learning rate to 4e-5. All other settings remain the same.

\subsection{Evaluation Settings}
\label{sec:evaluate}
We explicitly instruct all models to think step by step and to enclose the final answer within \texttt{\textbackslash boxed\{\}} \citep{hochlehnert2025soberlookprogresslanguage}.
We use the Math-Verify framework, a more robust way to extract and verify answers.

We first sample candidate responses in the direct-solving mode and then perform various test-time scaling strategies based on these candidate responses.
For base Qwen2.5-7B-Instruct, we configure the evaluation process to set the maximum new tokens of 4,096 and apply optimal hyperparameters.
For our GSR-7B, we configure the maximum new tokens of 32,768, temperature of 0.6 and top-p 0.95.
To mitigate the evaluation variance \citep{hochlehnert2025soberlookprogresslanguage}, we repeat 32 trials for every problem in AIME24 and AIME25, and 16 for all other benchmarks. 
We present the average accuracy across 16 or 32 samples generated directly for every problem as pass@1.

Then, we compute several TTS metrics based on non-overlapping groups of every four candidate responses: maj@4 (majority voting accuracy), BoN@4 (Best-of-N) and Ref/SelfRef@4. 
The final scores for these metrics are the average performance across all groups.
For all baseline models, we follow the exact hyperparameter values and the specific prompt recommended in their official documentation or model cards. For RRM and RM-R1, we employ a knockout tournament strategy \citep{guo2025rewardreasoningmodel}, a method of iterative pairwise comparison and elimination to determine the best answer, which effectively guides LLMs to perform BoN sampling.
Our method is evaluated with a maximum of 32,768 output tokens, 6,144 tokens for candidate response input (1,566 tokens per candidate response), temperature of 0.6 and top-p 0.95. 
We ensure that our method also runs 32 trials for AIME24 and AIME25, and 16 for the remaining benchmarks. 

\section{Dataset Curation}
\label{sec:dataset_curation}
For simplicity, we establish a data pipeline based on a single large-scale mathematics data set, OpenMathReasoning\citep{moshkov2025aimo}. OpenMathReasoning dataset is curated from the AoPS community forums, performed rigorous filtering, classification, transformation, and benchmark decontamination, containing 540K mathematical problems and 3.2M generations from DeepSeek-R1 and QwQ-32B.

The detailed procedure, including the numbers of problems and samples remaining at each stage, is documented in Table \ref{tab:pre_dataset_construction}.
In summary, the pipeline involves several filtering steps, such as removing generations not used in the Kaggle competition, removing problems without an extracted answer, and discarding any generations without a pass rate evaluated by Qwen2.5-72B-Instruct.
Subsequently, we aggregate the corresponding solutions for each problem, and then retaining problems with a pass rate between 0.25 and 0.9, yielding a prefiltered pool of 56K unique problems and 617K generations.

\begin{table*}[htbp]
\centering
\caption{Step by step filtering process for the construction of problems. The process starts from a large-scale raw dataset and applies a series of filtering  operations to yield the final curated dataset. The number of unique problems and their corresponding generated outputs are tracked at each step.}
\label{tab:pre_dataset_construction}
\resizebox{\linewidth}{!}{
\begin{tabular}{l l r r}
\toprule
\textbf{Step} & \textbf{Filtering Operation} & \textbf{Problems} & \textbf{Generations} \\
\midrule
- & Initial Raw Dataset & 540K & 3.2M \\
1 & Remove generations not used in Kaggle competition. & - & 2.2M \\
2 & Retain data with 'problem\_type' == 'has\_answer\_extracted'. & - & 1.3M \\
3 & Discard problems that have no 'pass\_rate\_72b\_tir'. & - & 1.2M \\
4 & Aggregate generations by their corresponding unique problem. & 116K & 1.2M \\
5 & Filter problems to keep only those with a 'pass\_rate\_72b\_tir' between 0.25 and 0.90. & 56K & 617K \\
\bottomrule
\end{tabular}%
}
\end{table*}

We construct the self-refinement dataset ($\mathcal{D}_{selfR}$) using these 56K problems. As summarized in Table \ref{tab:dataset_construction}, for each problem, we generate 6 candidate responses with Qwen2.5-7B-Instruct using temperature of 1.0, top-p of 0.95, and a maximum output length of 4096 tokens to foster diversity. Given the inherent difficulty of the problems, we implement a specific selection process for the 6 responses. We filter and construct a fixed size set of 4 candidate responses. This set is composed of all correct solutions from the initial candidate pool of 6, with the remaining filled by incorrect solutions to meet the required size of 4. The order of these 4 candidates is then randomized.
By combining each original problem with its set of candidate responses according to the prompt template in Appendix \ref{sec:prompt}, we constructed a dataset of 56K self-refinement problems. We further remove data where the prompt length exceeded 8,192 tokens, resulting in a final dataset of 53K samples.

\begin{table*}[htbp]
\centering
\caption{Step by step curation of final hybrid training datasets from the prefiltered pool.}
\label{tab:dataset_construction}
\resizebox{\linewidth}{!}{
\begin{tabular}{l l l r r}
\toprule
\textbf{Path} & \textbf{Step} & \textbf{Operation} & \textbf{Problems} & \textbf{Generations} \\
\midrule
- & - & Prefiltered Pool (from Table \ref{tab:pre_dataset_construction} Step 5). & 56K & 617K \\
$\mathcal{D}_{selfR}$ & 1 & Use all 56K problems for self-refinement generation. & 56K & - \\
& 2 & Generate 6 solutions as candidates using Qwen2.5-7B-Instruct. & 56K & 336K \\
& 3 & Construct augmented prompts for self-refinement. & 56K & - \\
& 4 & Remove data where the prompt length exceeded 8,192 tokens. & 53K & - \\
& 5 & Generate 10 solutions per problem using QwQ-32B. & 53K & 530K \\
& 6 & Subsample final solutions based on correctness. & - & 184K \\
$\mathcal{D}_{direct}$ & 1 & Randomly sample 184K generations from the pre-filtered pool. & - & 184K \\
Final Dataset & 1 & Merge the direct-answer dataset with self-refinement dataset. & - & 368K \\
\bottomrule
\end{tabular}%
}
\end{table*}

For these 53K self-refinement problems, we utilize QwQ-32B model and generate up to 10 solutions for each problem in our dataset. We use temperature of 0.7, top-p of 0.95, and limit generations to 16,384 tokens. 
The generated solutions by QwQ-32B are then filtered based only on the correctness. Specifically, if a problem yields between one and nine correct solutions (out of 10), we retain all of them. However, for problems where all 10 generated solutions are correct, we randomly sample four of them. This curation process yields our final training dataset $\mathcal{D}_{selfR}$ for self-refinement task of 184K samples.

For the direct-answer dataset, we randomly sample an equal number of 184K, the size of our self-refinement dataset, from the dataset obtained after Step 5 in Table \ref{tab:pre_dataset_construction} to create our $\mathcal{D}_{direct}$ dataset. We then merge the self-refinement dataset $\mathcal{D}_{selfR}$ with direct-answer dataset $\mathcal{D}_{direct}$ to create our final hybrid 368K training dataset.

\section{Case Study}
\label{sec:case_study}
In this section, we present a more detailed case study from AIME24. This case demonstrates that our method can recover from four incorrect candidate responses and produce a final correct solution.

\subsection{Original Question and Candidate Solutions}
The following is the original problem and four candidate responses provided to the model. Due to space constraints, the content of the candidate responses has been condensed.

\begin{tcolorbox}[
  breakable,
  title=Question and Candidate Solutions,
  colback=blue!5,
  colframe=blue!50!white,
  fonttitle=\bfseries,
  coltitle=white,
  colbacktitle=gray!70!blue,
  boxrule=0.5pt,
  arc=2mm,
]
You are an expert and creative solver, provided with a challenging problem and a set of candidate responses which may be correct, partially correct or even wrong.
\\
You should first fully summarize the connection between candidate responses and problem, then generate a new and superior solution. You should generate a correct solution yourself if all candidates are wrong. Don't copy candidates, use insights selectively and reason independently. \\
Problem:\\
Let $\triangle ABC$ have circumcenter $O$ and incenter $I$ with $\overline{IA}\perp\overline{OI}$, circumradius $13$, and inradius $6$. Find $AB\cdot AC$. Let's think step by step and output the final answer within \texttt{\textbackslash boxed\{\}}.
Candidate Response 1:\\
...Given the symmetry and the conditions, we can use the known result for the product of the sides in a right triangle:\\   
$$AB \cdot AC = 384$$
Thus, the product \(AB \cdot AC\) is:$$\boxed{384}$$
Candidate Response 2:\\
...Thus, the final answer is $$\boxed{26\sqrt{39}}$$
Candidate Response 3:\\
...Thus, the final answer is:$$\boxed{312}$$
Candidate Response 4:\\
...Thus, the final answer is:$$\boxed{26}$$
\end{tcolorbox}

\subsection{Thinking Process of Self-Refinement}
In this section, we present a more detailed trace of the model's thinking component and summary component, when provided with a problem and four candidate responses. 
We split and structure the model's raw thinking process into four parts, each marked by \textbf{a bold title}.
To provide analytical insights, we have inserted \textit{\textcolor{blue}{meta comment in blue italics}}, focusing on how the model analyzes, discerns, and selectively utilizes valuable information from four candidates responses. We give special attention to the model's ability to self-correct by analyzing candidate responses, identifying flaws and choosing an alternative decoding path. To make the model's direct engagement explicit, its verbatim references to the candidate responses are \textcolor{red}{highlighted in red}.

\begin{tcolorbox}[
  breakable,
  title=Thinking component of GSR-7B,
  colback=blue!5,
  colframe=blue!50!white,
  fonttitle=\bfseries,
  coltitle=white,
  colbacktitle=gray!70!blue,
  boxrule=0.5pt,
  arc=2mm,
]
\textbf{1. Initial analysis and calculation (Consistent with candidates)} 

Okay, let me try to tackle this problem step by step. ... First, I remember that there's a formula connecting the distance between the incenter and circumcenter ($OI$) with the circumradius ($R$) and inradius ($r$). The formula is $OI^2 = R(R - 2r)$. Let me plug in the given values: $R = 13$ and $r = 6$. So, $OI^2 = 13(13 - 12) = 13 \times 1 = 13$, so $OI = \sqrt{13}$. That's straightforward. Next, the problem states that $IA$ is perpendicular to $OI$. Since $IA \perp OI$, triangle $OIA$ is a right triangle with right angle at $I$. The hypotenuse here is $OA$, which is the circumradius, so $OA = 13$. Using the Pythagorean theorem on triangle $OIA$: $OA^2 = OI^2 + IA^2$. Plugging in the known values: $13^2 = (\sqrt{13})^2 + IA^2 \rightarrow 169 = 13 + IA^2 \rightarrow IA^2 = 156 \rightarrow IA = \sqrt{156}$. 

\textcolor{blue}{\textit{[Meta comment: The model's initial calculations align perfectly with the correct steps found in Candidates 1 and 2, thereby establishing a valid foundational starting point. The model also notes a critical flaw in Candidates 3 and 4: they employ an incorrect formula, $OI^2 = R^2 - 2Rr$. Although this formula coincidentally produces the correct numerical value for the specific given radius $R=13$, its structural inaccuracy leads the model to correctly flag these candidates as unreliable early in the process.]}} \vspace{1em} 

\textbf{2. Deep dive and critique of candidate strategies} 

\textcolor{blue}{\textit{[Meta comment on Candidate 1: The model observes that Candidate 1's final answer, 384, is presented without a clear derivation. The model formulates a hypothesis: this answer might originate from an unstated and potentially incorrect assumption that the triangle is right-angled at vertex A. It then proceeds to rigorously test this hypothesis.]}} \vspace{1em} 

Alternatively, perhaps the triangle is right-angled? If $\triangle ABC$ is right-angled, then the circumradius would be half the hypotenuse. Since the circumradius is 13, then the hypotenuse would be 26. Suppose angle A is $90^\circ$, then BC would be the hypotenuse, so $BC=26$. Then, using the inradius formula: $r = (AB + AC - BC)/2$. Given $r=6$, so $(AB + AC - 26)/2 = 6 \rightarrow AB + AC = 38$. Also, by Pythagoras, $AB^2 + AC^2 = 26^2 = 676$. ... $2 AB \cdot AC = 1444 - 676 = 768 \rightarrow AB \cdot AC = 384$. 
\textcolor{red}{Wait, that's one of the candidate answers (Candidate 1's answer is 384). But does this satisfy the condition $IA \perp OI$?} ... Let me set coordinates: Let’s place $A$ at $(0,0)$, $B$ at $(0,b)$, $C$ at $(c,0)$. ... in a right-angled triangle at A(0,0), with legs on axes, ... the incenter is at $(r, r)$. So here, I would be at $(6,6)$. The circumcenter $O$ is the midpoint of $BC$, which is at $(c/2, b/2)$. Now, we need to check if $IA$ is perpendicular to $OI$. Vector $\vec{IA}$ is from I(6,6) to A(0,0): $(-6, -6)$. Vector $\vec{OI}$ is from O(c/2, b/2) to I(6,6): $(6 - c/2, 6 - b/2)$. Their dot product should be zero: $(-6)(6 - c/2) + (-6)(6 - b/2) = 0 \rightarrow -36 + 3c/2 - 36 + 3b/2 = 0 \rightarrow -72 + (3/2)(b+c) = 0$. But since $b+c = 38, \rightarrow -72 + (3/2)*38 = -72 + 57 = -15 \neq 0$. Thus, the dot product is not zero, so the condition $IA \perp OI$ is not satisfied.
...\textcolor{red}{but the calculation gave AB·AC=384, which is the first candidate's answer, but it doesn't satisfy the perpendicular condition.} So that must be incorrect... Therefore, the triangle cannot be right-angled at A. Hmm, so that approach was wrong. 

\vspace{1em} \textcolor{blue}{\textit{[Meta comment on Candidate 2: The model shifts its analysis to the other candidates and identifies a valuable, correct intermediate step within Candidate 2's reasoning. It decides to pursue this line of reasoning to evaluate its potential and identify where, if at all, it deviates from a correct solution path.]}} \vspace{1em}

...Wait, another formula for $IA$ is $IA = 2R \sin(A/2)$. Let me check that. Yes, I think that's correct. ... $IA = 2R \sin(A/2) \rightarrow \sqrt{156} = 2 \cdot 13 \cdot \sin(A/2) \rightarrow \sin(A/2) = \sqrt{156}/(26) = \sqrt{39}/13$. ... Thus, $\sin A = 2 \sin(A/2) \cos(A/2) = 2 \cdot (\sqrt{39}/13) \cdot (\sqrt{130}/13) = \dots = (2\sqrt{30})/13$. ... We had $\sin A = 2 \sqrt{30}/13$. Then $\cos A = \sqrt{1 - \sin^2 A} = \sqrt{1 - (4 \cdot 30)/169} = \sqrt{(169 - 120)/169} = \sqrt{49/169} = 7/13$. 

\vspace{1em}  \textcolor{blue}{\textit{[Meta comment: The model confirms the validity of Candidate 2's calculations regarding the trigonometric properties of angle A, acknowledging this as a key insight. However, it also diagnoses the critical failure point in Candidate 2's approach: the strategy becomes untenable when it hinges on calculating the semiperimeter, 's', a step where the candidate's method falters. The model elects to adopt the correct trigonometric values derived by Candidate 2 while consciously avoiding the flawed semiperimeter-dependent strategy.]}}  \vspace{1em}

...So, $6s = (1/2) \cdot AB \cdot AC \cdot (2\sqrt{30}/13) \rightarrow 6s = (AB \cdot AC \cdot \sqrt{30})/13 \rightarrow AB \cdot AC = (6s \cdot 13)/\sqrt{30}$. Hmm, so if I can find $s$, then I can compute $AB \cdot AC$...
\textcolor{red}{This is the path Candidate 2 took, but they failed to find $s$ correctly.} I will retain the trigonometric values for angle A but seek a more direct path to the solution, avoiding this semiperimeter dead end. \vspace{1em} 

\textbf{3. Synthesizing a novel, correct solution from useful parts} 

\textcolor{blue}{\textit{[Meta comment: After systematically identifying the flaws in Candidates 1, 3, and 4, and the dead-end in Candidate 2's approach, the model determines that a novel solution path is required. The chosen strategy is a synthesis: it leverages the correct trigonometric values for angle A, which were validated from Candidate 2's work, and integrates them with a powerful trigonometric identity not utilized by any of the candidates.]}} \vspace{1em} 

...We have $r=6, R=13$, so: $6 = 4 \cdot 13 \cdot \sin(A/2) \sin(B/2) \sin(C/2)$. We already found $\sin(A/2) = \sqrt{39}/13$. Let me plug that in: $6 = 52 \cdot (\sqrt{39}/13) \cdot \sin(B/2) \sin(C/2) \rightarrow 6 = 4\sqrt{39} \cdot \sin(B/2) \sin(C/2) \rightarrow \sin(B/2) \sin(C/2) = 6/(4\sqrt{39}) = 3/(2\sqrt{39})$. ... Also, since $B + C = \pi - A$, $\sin(B/2)\sin(C/2)= [\cos((B - C)/2) - \cos((B + C)/2)]/2 = [\cos((B - C)/2) - \sin(A/2)]/2$. So, $[\cos((B - C)/2) - \sqrt{39}/13]/2 = 3/(2\sqrt{39}) \rightarrow \cos((B - C)/2) - \sqrt{39}/13 = 3/\sqrt{39} \rightarrow \cos((B - C)/2) = 3/\sqrt{39} + \sqrt{39}/13 = \dots = 2\sqrt{39}/13$. ... Now, using the identity for $\sin B \sin C$: $\sin B \sin C = [\cos(B - C) - \cos(B + C)] / 2$. We have $\cos(B - C) = 2 \cos^2((B - C)/2) - 1 = 2 \cdot ((2\sqrt{39}/13)^2) - 1 = \dots = 143/169$. And $\cos(B + C) = -\cos A = -7/13$. $\sin B \sin C = [143/169 - (-7/13)] / 2 = [143/169 + 91/169] / 2 = (234/169) / 2 = 117/169$. ... Thus, $AB \cdot AC = 4R^2 \sin B \sin C = 4 \cdot (13)^2 \cdot (117/169) = 4 \cdot 169 \cdot (117/169) = 4 \cdot 117 = 468$. \vspace{1em} 

\textbf{4. Final verification of the self-derived answer} 

...Hmm, so the correct answer should be 468. \textcolor{red}{But none of the candidates have this.} ... let me see if that makes sense with the area. The area $\Delta = (AB \cdot AC \sin A)/2 = (468 \cdot (2\sqrt{30}/13))/2 = (468 \cdot \sqrt{30})/13 = 36\sqrt{30}$. Also, the area is $rs = 6s$. Thus, $6s = 36\sqrt{30} \rightarrow s = 6\sqrt{30}$. ... The sides are $a = 4\sqrt{30}$, and $b + c = 2s - a = 12\sqrt{30} - 4\sqrt{30} = 8\sqrt{30}$. Thus, $b+c = 8\sqrt{30}$, and $bc=468$. ... We can check if these satisfy the Law of Cosines: $\cos A = (b^2 + c^2 - a^2)/(2bc)$. We have $\cos A = 7/13$. ... $b^2 + c^2 = (b+c)^2 - 2bc = (8\sqrt{30})^2 - 2(468) = 1920 - 936 = 984$. Plugging into Law of Cosines: $(984 - 480)/(2 \cdot 468) = 504/936 = 7/13$. This matches the earlier result. Thus, the calculations are consistent. Therefore, the correct answer should be 468. 
\end{tcolorbox}

\subsection{Summary Component of Self-Refinement}
We now present the summary component generated by the model. This component serves as a concise distillation of the core content detailed in the preceding chain-of-thought analysis. It is worthy that the model autonomously provides a novel solution path, distinct from all candidate responses, to arrive at the correct answer of \textbf{468}.

\begin{tcolorbox}[
  breakable,
  title=Summary component of GSR-7B,
  colback=blue!5,
  colframe=blue!50!white,
  fonttitle=\bfseries,
  coltitle=white,
  colbacktitle=gray!70!blue,
  boxrule=0.5pt,
  arc=2mm,
]
To solve the problem, we need to find the product \(AB \cdot AC\) for triangle \(ABC\) with circumradius \(R = 13\), inradius \(r = 6\), and the condition that \( \overline{IA} \perp \overline{OI} \). \\Key Steps:\\1. **Distance Between Incenter and Circumcenter**:   Using Euler's formula:   
\begin{align*}
OI^2 =& R(R - 2r) = 13(13 - 12) = 13\\ 
&\implies OI = \sqrt{13}   
\end{align*}
2. **Right Triangle Condition**:   Since \( \overline{IA} \perp \overline{OI} \), triangle \(OIA\) is right-angled at \(I\). Using the Pythagorean theorem:   
\begin{multline*}
OA^2 = OI^2 + IA^2 \implies 13^2 = (\sqrt{13})^2 + \\ IA^2 \implies IA^2 = 156 \implies IA = \sqrt{156}   
\end{multline*}
3. **Trigonometric Relationships**:   - The distance \(IA\) can also be expressed as \( IA = 2R \sin\left(\frac{A}{2}\right) \):    \begin{align*}
& \sqrt{156} = 2 \cdot 13 \cdot \sin\left(\frac{A}{2}\right) \\  & \implies \sin\left(\frac{A}{2}\right) = \frac{\sqrt{39}}{13}    
\end{align*}
- Using the double-angle identity for \(\sin A\): 
\begin{align*}
\sin A & = 2 \sin\left(\frac{A}{2}\right) \cos\left(\frac{A}{2}\right) \\
&= 2 \cdot \frac{\sqrt{39}}{13} \cdot \sqrt{\frac{130}{169}} = \frac{2\sqrt{30}}{13}
\end{align*}
- The cosine of angle \(A\) is:     \[     \cos A = \sqrt{1 - \sin^2 A} = \frac{7}{13}     \]4. **Product of Sides Using Law of Sines**:   - The sides \(AB\) and \(AC\) are given by:     \[     AB = 2R \sin C \quad \text{and} \quad AC = 2R \sin B     \]   - The product \(AB \cdot AC\) is:     \[     AB \cdot AC = 4R^2 \sin B \sin C     \]   - Using trigonometric identities and the given conditions, we find:     \[     \sin B \sin C = \frac{117}{169}     \]   - Substituting \(R = 13\) and \(\sin B \sin C\):     \[     AB \cdot AC = 4 \cdot 13^2 \cdot \frac{117}{169} = 4 \cdot 117 = 468     \] Final Answer:\[\boxed{468}\]
\end{tcolorbox}

\end{document}